\documentclass[
a4paper,11pt,twoside,openright
]{report}

\usepackage{amsmath}
\usepackage{tabularx}
\usepackage{hyperref}
\usepackage[refpages]{gloss}
\usepackage[dvips]{graphicx}
\usepackage{rotating}
\usepackage{titlesec}
\usepackage{url}
\usepackage{fancyhdr}
\usepackage{latexsym}
\usepackage[chapter]{algorithm}
\usepackage{algorithmic}

 \hypersetup{
   pdftitle = {Thematic Annotation: extracting concepts out of documents.},
   pdfsubject = {Technical report EPFL/LIA, ID: IC/2004/68},
   pdfkeywords = {technical, repOrt, epfl, andrews, topic extraction,
   keyword extraction, segmentation, data mining, computational
   linguistic, ontology, semantic database, semantic},
   pdfauthor = {\textcopyright\ Pierre Andrews},
   pdfcreator = {\LaTeX\ with Emacs on OS X},
 } 

\newcommand{\clearemptydoublepage}
 {\newpage{\pagestyle{empty}\cleardoublepage}}

\renewcommand{\glosspage}[1]{\ \dotfill\mbox{\ p.#1}} 
\setglosstext{none}{}
\makegloss %

\fancypagestyle{plain}{%
  \fancyhead{} %
}

\titleformat{\section}
{\normalfont\Large\sffamily}
{\filright{\textbf{\thesection}}}{1em}{\filright\textbf}
                                                                                                                          
\titleformat{\subsection}
{\normalfont\normalsize\sffamily}
{\filright{\textbf{\thesubsection}}}{1.3em}{\filright\textbf}
 
\titleformat{\subsubsection}
{\normalfont\normalsize\sffamily}
{}{0em}{\filright\textbf}

\begin{document}

\begin{titlepage}

  \setcounter{page}{1}
  
  \vspace*{2cm}
  \begin{flushright}
    {\sc{\normalfont\Huge\sffamily{\textbf{Thematic Annotation:}}}}
    \\[0.2cm]
    {\sc{\normalfont\Huge\sffamily{\textbf{extracting concepts out of documents}}}}
  \end{flushright}
  \vspace*{1cm}
  \begin{flushright}
    {\large{\sc{Pierre Andrews}}}
    \\
    {\footnotesize{\url{pierre.andrews@a3.epfl.ch}}}
    \\
    {\large{\sc{Martin Rajman}}}
    \\
    {\footnotesize{\url{martin.rajman@epfl.ch}}}
  \end{flushright}
  \vfill
  \begin{flushright}
    \hfill {\parbox{8cm}{\footnotesize{\textbf{Technical Report IC/2004/68}}}}
    \\[0.4cm]
    {\parbox{8cm}{\footnotesize{School of Computer \& Communication Sciences\\
          \textbf{Swiss Federal Institute of Technology, Lausanne}}}}
    \\[0.4cm]
    {\parbox{8cm}{\footnotesize{\textbf{Artificial Intelligence Laboratory}\\
          Institute of Core Computing Science}}}
    \\[0.4cm]
    {\parbox{8cm}{\footnotesize{August 2004}}}
  \end{flushright}
\end{titlepage}
\clearemptydoublepage

\begin{abstract}
  Semantic document annotation may be useful for many tasks. In
  particular, in the framework of the MDM
  project\footnote{http://www.issco.unige.ch/projects/im2/mdm/},
  topical annotation -- i.e. the annotation of document segments with
  tags identifying the topics discussed in the segments -- is used to
  enhance the retrieval of multimodal meeting records. Indeed, with
  such an annotation, meeting retrieval can integrate topics in the
  search criteria offered to the users.
  
  Contrarily to standard approaches to topic annotation, the technique
  used in this work does not centraly rely on some sort of -- possibly
  statistical -- keyword extraction. In fact, the proposed annotation
  algorithm uses a large scale semantic database -- the EDR Electronic
  Dictionary\footnote{http://www.iijnet.or.jp/edr/} -- that provides a
  concept hierarchy based on hyponym and hypernym relations.This
  concept hierarchy is used to generate a synthetic representation of
  the document by aggregating the words present in topically
  homogeneous document segments into a set of concepts best preserving
  the document's content.
  
  The identification of the topically homogeneous segments -- often
  called Text Tiling -- is performed to ease the computation as the
  algorithm will work on smaller text fragments. In addition, it is
  believed to improve the precision of the extraction as it is
  performed on topically homogeneous segments. For this task, a
  standard techniques -- proposed by \cite{hearst94multiparagraph} --
  relying on similarity computation based on vector space
  representations have been implemented. Hence, the main challenge in
  the project was to create a novel topic identification algorithm,
  based on the available semantic resource, that produces good results
  when applied on the automatically generated segments.
  
  This new extraction technique uses an unexplored approach to topic
  selection. Instead of using semantic similarity measures based on a
  semantic resource, the later is processed to extract the part of the
  conceptual hierarchy relevant to the document content. Then this
  conceptual hierarchy is searched to extract the most relevant set
  of concepts to represent the topics discussed in the document.
  Notice that this algorithm is able to extract generic concepts that
  are not directly present in the document.

  The segmentation algorithm was evaluated on the Reuters corpus,
  composed of 806'791 news items. These items were aggregated to
  construct a single virtual document where the algorithm had to detect
  boundaries. These automatically generated segments were then
  compared to the initial news items and a metric has been developed
  to evaluate the accuracy of the algorithm.
  
  The proposed approach for topic extraction was experimentally tested
  and evaluated on a database of 238 documents corresponding to
  bibliographic descriptions extracted from the INSPEC
  database\footnote{http://www.iee.org/publish/inspec/}. A novel
  evaluation metric was designed to take into account the fact that
  the topics associated with the INSPEC descriptions -- taken as the
  golden truth for the evaluation -- were not produced based
  on the EDR dictionary, and therefore needed to be approximated by
  the available EDR entries.
  
  Alltogether, the combination of existing document segmentation
  methods -- i.e text tiling -- with novel topic identification ones
  leads to an additional document annotation useful for more robust
  retrieval.

\end{abstract}

\setcounter{page}{4}

\thispagestyle{empty}
\tableofcontents
\thispagestyle{empty}

\chapter{Introduction}
\label{chap:introduction}

\section{Goals and issues}
\label{sec:intro:goals-issues}

Automatic document summarization is often regarded as a technique to
extract a set of sentences from the document. This set of sentences
can then be displayed to the user for an easier understanding of a
document's themes. However, when used for information retrieval, theme
descriptors should be less constraining than full sentences.  The user
might prefer to describe his query with simple keywords and keyword
extraction methods have been developed for this purpose.  The problem
with such a method is that it's hard to retain the
\gloss[word]{semantic}s of the whole document in a \emph{few
  keywords}.  Hence, an annotation method offering a tradeoff between
summarization -- that preserves most of the semantics -- and keywords
extraction -- that are easier to query, but loose a lot of the
semantics -- has been investigated.

The aim of this project is to provide a list of topics -- i.e. a small
list of words or \gloss[word]{compounds} each representing a concept--
for the indexation of the processed documents. Hence a technique
extracting a \emph{limited} list of topics that represents the
document subjects has been developed.  The main goal is then, not to
extract keywords from the documents, but words representing the topics
present in this document. These words should preserve more information
about the content of the document as they describe concepts and not just
words present in this document.

We believe that the use of an \gloss[word]{ontology} can help the for
this process, in particular for aggregation of conceptually similar
words. A \gloss[word]{semantic} database (like the EDR Electronic
Dictionary) provides a dictionary where all words are associated with
a gloss and the concepts it can represent. Each of these concepts is
also positioned in a \gloss[word]{DAG} representing a
\gloss[word]{semantic} classification; each concept is attached to one
or more super-concepts -- having a more general meaning -- and zero or
more sub-concepts -- having a more precise meaning.  The EDR database
can be used to identify the concepts attached to each word of our
document. The document is then represented as a bag of concepts, and
the challenge is to use the conceptual structure to deduce the discussed
topics.

The EDR database provides about 489'000 concepts in its hierarchy;
such a large number implies that, for one word in our document, we
will usually have more than one concept associated with it.
Aggregating these concepts for a big document can be computationally
inefficient as the resulting conceptual \gloss[short]{DAG} will be
large as well. For this reason, the direct processing of large
documents is not realistic in the perspective of the extraction of
relevant topics.

An possible approach to process large documents is to arbitrarily
divide them into multiple parts and to extract the topics for each of
these parts.  However, to keep a certain coherence in the segmentation
and to give the topic extraction algorithm a better chance, the
segmentation cannot be made randomly. Keeping a homogeneity in the
segment's \gloss[none]{semantic}semantics increases the probabilities
to extract the more relevant concepts from each part of the document.
Good techniques already exist to segment texts in homogeneous blocks
and provide simple statistical algorithms that can be implemented
without many computational obstacles.

\section{Segmentation}
\label{sec:intro:segmentation}

The Text Tiling method \cite{hearst94multiparagraph} that has been
chosen is a simple \gloss[word]{statmeth} (see chapter
\ref{cha:extraction}). It uses a \gloss[word]{bow} representation of
the document to identify segments:
\begin{itemize}
\item the document is split in N windows of fixed size,
\item a \gloss[word]{bow} representation is constructed associating a
  weight to each token,
\item the distance between every two consecutive window is computed,
\item thematic shifts are identified by searching the positions in the
  text where these distances are minimal.
\end{itemize}

This method has the advantage of requiring little preprocessing and no
external knowledge source. Segments are only bound by intra document
computations. However, if not tuned correctly this technique can lead to
poor results. Initial windows size is critical and varies for each
text. User input would be optimal to adjust the algorithm parameters
, but is not in the scope of our system.  Luckily, to perform the
topic extraction, the segments do not have to be exact and a
\emph{noisy} segmentation is therefore admissible in our case.

\section{Topic Extraction}
\label{sec:intro:topic-extraction}

The topic extraction method that has been developed for this project
has barely been studied from such a point of view before. Indeed the
use of an external knowledge source often translates into additions to
the basic \gloss[word]{bow} representation produced for the document,
or by the computation of \gloss[word]{semantic} similarities (see
section \ref{cha:state-art}).  In this project, another approach has
been chosen:
\begin{itemize}
\item for each word of the document, we identify the
  \gloss[none]{lemma}lemmas from which it can derive,
\item a list of leaf concepts is extracted from the semantic database for each
  \gloss[word]{lemma},
\item starting from the leaf level, the conceptual hierarchy is
  constructed level by level,
\item a set of concepts corresponding a cut\footnote{see section
    \ref{sec:cut-extraction} for the definition of a cut in a
    \gloss[short]{DAG}.} in this hierarchy is selected to describe the
  document.
\end{itemize}

This method is believed to be more efficient as it does not compute
the similarity of each possible pair of concepts but only constructs a
subhierarchy of the complete ontology. Preprocessing for this task
have been kept simple and only part of speech disambiguation is
performed (see section \ref{sec:ext:preprocessing}) before the
construction of the \gloss[short]{DAG}. The weak point of the current
version of the algorithm is the absence of any Word Sense
Disambiguation (see section \ref{sec:ext:disambiguation}), as a word
in the document can trigger more than one concept in the hierarchy --
each of these concepts representing a different word sense.

\section{Document Structure}
\label{sec:intro:document-structure}

This report has been divided in four main parts. Chapter
\ref{cha:state-art} presents the state of the art of the existing
techniques that have influenced the developments made during the
project. The next Chapter (Chapter \ref{cha:edr-dict-ontol}) presents
the main resource used during this project and the constraints that
had to be taken into account to conform to the available linguistic
resource (in addition, see Section \ref{sec:ext:token}).

The two main parts of the project are presented in Chapters
\ref{cha:segmentation} and \ref{cha:extraction}. The first presents
the implementation of the Text Tiling techniques, while the second
describes our novel topic extraction technique.

Chapters \ref{cha:evaluation} and \ref{cha:conclusion} present the
evaluation process and the conclusions that were reached from
it. \ref{cha:future-work} offers some hints on the future work that
can be lead in continuation to this project.

\chapter{State Of the Art}
\label{cha:state-art}

\section{Segmentation}
\label{sec:state-art:segmentation}

\subsection*{Text Tiling}

Automatic textual segmentation is a technique that has been well
studied and developed for several years. M.Hearst
\cite{hearst94multiparagraph} introduced a statistical method using a
\gloss[word]{bow} to segment text. In his work, segmentation was
based on \emph{geometric} distance between vectorial representations of
 text. M.Hearst demonstrated that only simple preprocessing was
needed to perform well and used only a \gloss[word]{stoplist}. This
method has been extended in a few ways to perform better or to be more
robust.

\subsection*{Extended Geometric Methods}

N.Masson \cite{ferret98thematic} proposed to extend the notion of distance
computation by using lexical information. A lexical network
containing relations between words is used to extend the initial
vectorial representation of the document. Hence, when a word A is
present, a neighbour word B (not in the document) can be introduced in
the \gloss[word]{bow} to render the lexical proximity that exists
between them.  The use of such an external resource while improving
the technique's robustness introduces language dependency.  K.Richmond
\cite{richmond97detecting} develops a language independent
computation of distances in text. His idea is to exclude
non-content words from the vectorial representation; the use of the
content words' distribution pattern in documents provides interesting
information to detect these content bearing words.

\subsection*{Lexical Chains}

Another totally different approach to segmentation uses
\gloss[word]{lexicalchains} (see \cite{okumura94word} and
\cite{popescu04shallow}) to detect subject boundaries. Chains
containing sibling words are constructed while traversing the text.
Each time a word is considered, it is added to a chain to keep the
best lexical cohesion in each of them. Lexical chains are well suited
for this purpose as they keep information on the actual context and
help to disambiguate when more than one lexical possibility exists for
a word. Once the chains are constructed, it is quite easy to find the
segment boundaries.  At each position in the text, the number of
starting chains and the number of ending chains are computed. These
measurements give the plausibility of a segmentation at each position
of the text and can be used to select the \emph{right} ones.

All these techniques have proved to be reliable. However, some of them
are more or less robust or sometimes require non-obvious
implementations.

\section{Topic Extraction}
\label{sec:state-art:topic-extraction}

Keyword extraction is often used in documents indexing, which is a key
tool for Internet searches and is therefore well developed. A frequent
technique is to use the \gloss[word]{TFIDF} weighting
\cite{Salton88Term} of the \gloss[word]{bow} to select the content
bearing words. This technique has been extended in many ways.

\subsection*{\gloss[Long]{TDT}}

Topic detection is an extended keyword extraction technique mostly
used in \gloss[short]{TDT} \cite{allan98TDT} where topics are tracked
in a news stream, each \emph{new} topic is marked and the following
news are attached to the corresponding topic. The language models that
have been developed for this purpose (see \cite{nallapati03TDT} for
example) are not really in the scope of this paper, but demonstrate
how \gloss[word]{semantic} and lexical information can be used to
improve the processing.

\subsection*{Keyword Extraction}

Keyword extraction, as previously discussed, only extracts words
present in the document. This is sometimes not the desired behaviour
as a word alone, out of its context, does not retain as much
information.  Therefore, keyword extraction can be disappointing for
the user when it only presents a few words, whereas it is still good
for indexing as the number of keywords does not have to be as
restricted for this purpose. A vector extension technique like the one
presented by \cite{ferret98thematic} could provide keywords more
descriptive of the whole document even out of their context.

\subsection*{Lexical Chains}

Barzilay \cite{barzilay97using} proposes to use
\gloss[word]{lexicalchains} for summarization. Constructing a set of
\gloss[word]{lexicalchains} for a document grasps context and lexical cohesion.
Barzilay uses these chains to extract important sentences from the
text. However, we could think of an algorithm using these chains to
extract keywords: for the best chain constructed from the document,
one or more words representing the information contained in this chain
can be selected.

\subsection*{Semantic Similarity}

These techniques use external knowledge sources (i.e.  lexical
networks or ontologies) to compute the lexical cohesion of words.
Computing the cohesion between words is a widely open subject, as
there are many ways of doing it. Lexicographic distance is an easy way
to compute a measure but only represents word cohesion in terms of
number of shared letters -- which is not always a criterion of
\gloss[word]{semantic} closeness. When semantic databases are used to
compute the cohesion, a few \emph{formulas} provide metrics for the
\gloss[word]{sd} or similarity of two words.
\cite{budanitsky01semantic} gives a good survey on the existing
metrics.

A project developed in parallel to this one \cite{plas04extraction}
uses Semantic Similarity for keyword extraction. Its goal is to
extract keywords present in the text by using, in addition to the
relative frequency ratio (RFR) filtering, the semantic context
described by the semantic distance between consecutive words.

Unlike our project, there is no will to extract generic concepts
higher in the hierarchy. This one is only used to compute the semantic
similarity between words. Then, the similarity value is used to
create a hierarchical clustering of all the words in the documents. The
best clusters are then selected and a representative keyword in each
cluster is selected to annotate the document.

\chapter{Semantic Resource}
\label{cha:edr-dict-ontol}

\section{Description}
\label{sec:edr:about}

The EDR Electronic Dictionary \cite{yokoi95edr} is a set of resources
developed by the Japan Electronic Dictionary Research Institute, Ltd.
and maintained by the Communications Research Laboratory
(CRL)\footnote{http://www.crl.go.jp/overview/index.html} that can be
used for natural language processing. It is composed by:
\begin{itemize}
\item an English and a Japanese word dictionary,
\item a Concept dictionary,
\item an English $\leftrightarrow$ Japanese bilingual dictionary,
\item an English and a Japanese co-occurrence dictionary,
\item an English and a Japanese corpus.
\end{itemize}

This electronic resource is consequent; it contains 420'115 dictionary
entries for 242'493 unique word forms. Compared to the
WordNet\footnote{WordNet is one of the main ontology freely available
  and is often used in Natural Language Processing researches.
  www.cogsci.princeton.edu/\~wn/} ontology that contains 203'145 total
word senses, EDR contains 488'732 concepts and 507'665 simple
\gloss[word]{hypernym}/\gloss[word]{hyponym} links between them.

For the current project, the English and the Concept dictionary were
mostly used.  The English dictionary provides information on basic and
common words:
\begin{itemize}
\item \gloss[word]{lemma} (called \emph{HeadWord}),
\item pronunciation,
\item grammatical information such as inflection patterns and
  \gloss[word]{pos} tags,
\item \gloss[word]{semantic} information that links to the Concept
  dictionary.
\end{itemize}
Its main purpose is to make a correspondence between English words and
the conceptual dictionary and to provide basic grammatical
information. The Concept dictionary can be considered as an
\gloss[word]{ontology}, as it represents the
\gloss[word]{hypernym}/\gloss[word]{hyponym} links between common
concepts. This dictionary does not make distinctions between English
and Japanese concepts, as there is no need to think that there is one.
Its content is divided in three sub-dictionaries:
\begin{itemize}
\item the \emph{HeadConcept} dictionary which provides information on
  the human meaning of each concept with a gloss or a representative
  word,
\item the classification dictionary, placing each concept in the
  conceptual hierarchy,
\item the description dictionary, containing additional links between
  concepts that cannot be described by the super/sub classification.
\end{itemize}
The EDR dictionary is a complex structure with a considerable amount
of information that has not been cited here. Reference to the complete
documentation for further information is strongly advised
\cite{edr95}. An example of dictionary entry is presented in table
\ref{tab:edr:ewdextract}.

\begin{table}[htb]
\begin{tabularx}{\textwidth}{|c|X|X|c|X|X|X|}
  \hline
  \textbf{Type}&\textbf{word}&\textbf{syllable}&\textbf{pron.}&\textbf{POS}&\textbf{word
    form}&\textbf{flex.}\\
  \hline \hline
  S&Cypriot&Cyp ri ot&s'ipriXet&Com. Noun&Sing.&\\
  \hline
  S&Cypriot&Cyp ri ot&s'ipriXet&Adj.&Pos. Deg.&\\
  \hline
  S&Cypriote&Cyp ri ote&s'ipriXet&Com. Noun&Sing.&"s" "-" "s"\\
  \hline
  S&Cypriote&Cyp ri ote&s'ipriXet&Adj.&Pos. Deg.&\\
  \hline
  S&Cyprus&Cy prus&s'aiprXes&Prop. Noun&Sing.&\\
  \hline
  S&Cyprus&Cy prus&s'aiprXes&Prop. Noun&Sing.&\\
  \hline
  S&Cyrano de Bergerac&Cy ra no de Ber ge rac&...&Prop. Noun&Sing.&\\
  \hline
\end{tabularx}
\begin{tabularx}{\textwidth}{|c|X|c|c|}
  \hline
  \textbf{gloss}&$F_{word}$&$F_{word}^{concept}$\\
  \hline \hline
  a Greek dialect used in Cyprus&3&0\\
  \hline
  pertaining to Cyprus&1&1\\
  \hline
  -&0&0\\
  \hline
  -&0&0\\
  \hline
  an island in the Mediterranean sea named Cyprus&4&2\\
  \hline
  a country called Cyprus&4&2\\
  \hline
  the main character in the play Cyrano de Bergerac&0&0\\
  \hline
\end{tabularx}
\caption{Extract of the English Word dictionary}
\label{tab:edr:ewdextract}
\end{table}

\begin{figure}[htbp]
  \centering \includegraphics[width=\textwidth]{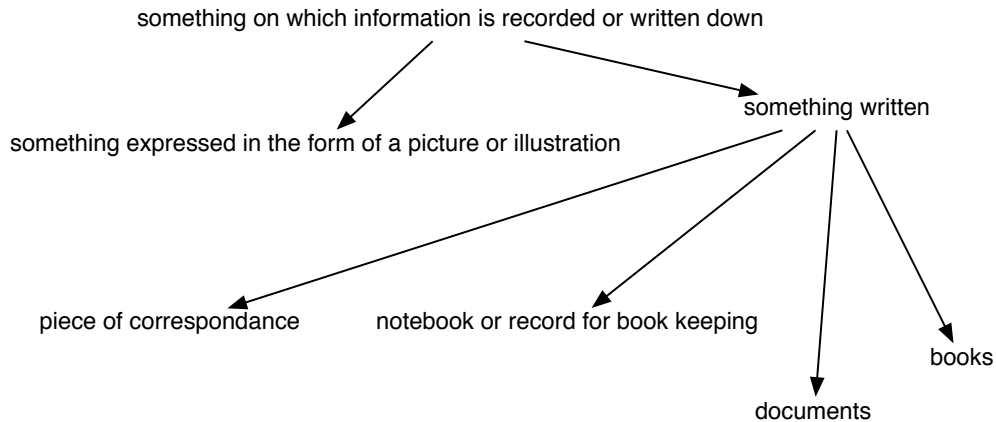}
  \caption{Extract of the concept classification}
  \label{fig:edr:extract}
\end{figure}

Even if figure \ref{fig:edr:extract} presents a simple tree structure
extracted from the concept dictionary, it is more frequent to
encounter complex structures where \gloss[none]{hyponym}hyponyms have
more than one \gloss[word]{hypernym} -- due to multiple inheritance --
and that cannot be considered as trees. Sometimes, in this document,
the conceptual hierarchy will be referred to as the:
``\gloss[word]{DAG}''.

This data is provided in an structured textual form (see figure
\ref{fig:encoded1} and \ref{fig:encoded2}). For ease of use, it has
been processed and entered in a database preserving the data structure
present in the original format. This processing has been performed in
an earlier project at the LIA\footnote{Artificial Intelligence
  Laboratory.  http://liawww.epfl.ch} and facilitates the access to
the data.  Its major strength is the access
\gloss[short]{API}, a feature provided by the querying capacities of
any modern database, which makes the development of different
applications around EDR faster. A description of the major tables used
during the project can be found in section
\ref{sec:edr:interesting-tables}.

\begin{figure}[htbp]
\centering
\begin{verbatim}
<Record Number>                         EWD1364642
<Headword Information>
        <Headword>                      supply
        <Invariable Portion of Headword and Adjacency Attributes Pair>
                                        suppl(Verb with Initial 
                                        Consonant Sound, Invariable 
                                        Portion of Verb Headword -
                                        Inflection Pattern y)
        <Syllable Division>             sup/ply
        <Pronunciation>                 sXepl'ai
<Grammar Information>
        <Part of Speech>                Verb
        <Syntactic Tree>
        <Word Form and Inflection Information>
        <Word Form Information>         Invariable Portion of Verb
        <Inflection Information>        Inflection Pattern y
        <Grammatical Attributes>
        <Sentence Pattern Information>  Must take a direct object 
                                       (direct object is a 
                                        phrase);Takes a prepositional
                                        phrase beginning with the 
                                        preposition 'to'
        <Function and Position Information>
        <Function Word Information>
<Semantic Information>
        <Concept Identifier>            0ec944
        <Headconcept>
                <Japanese Headconcept>  支給する［シキュウ・スル］
                <English Headconcept>   supply
        <Concept Explication>
                <Japanese Concept Explication>
                                        物をあてがう
                <English Concept Explication>
                                        to supply goods
<Pragmatic and Supplementary Information>
        <Usage>
        <Frequency>                     122/234
<Management Information>
        <Management History Record>     3/4/93
\end{verbatim}
\caption{Example of English Word Dictionary Records (Verb)}
\label{fig:encoded1}
\end{figure}
\begin{figure}[htbp]
\centering
\begin{verbatim}
<Record Number>                         CPH0314159
<Concept Identifier>                    3d0ecb
<Headconcept>
        <English Headconcept>           borrow
        <Japanese Headconcept>          借りる[カリ・ル]
<Concept Explication>
        <English Concept Explication>   to use a person's property after
                                        promising to ...
        <Japanese Concept Explication>  返す約束で,他人のものを使う
<Management Information>
        <Management History Record>     Date of record update "93/04/26"
\end{verbatim}
\caption{Example of Headconcept Records}
\label{fig:encoded2}
\end{figure}

\section{Version Issues}
\label{sec:edr:version-issues}

Computerized linguistic resources are rarely as perfect as they should
be. Annotating and defining a large quantity of linguistic data
requires mostly human work and ``error is human''. During this
project, the use of the EDR database revealed strange values for some
of its elements. However it has not been checked whether these errors
were in the original data or came from the text to database
conversion.

\subsection*{Anonymous Concepts}
\label{sec:edr:anonymous-concepts}
The documentation of EDR is clear on the fact that, for each
\emph{headconcept} present in the hierarchy, there should be at least
a gloss -- the \emph{headword} is not a must -- explaining
the concept. Intuitively this seems right: how could the EDR creator
have the knowledge to insert a concept (i.e. they know that this
  concept exists) without being able to define it?

Since a great part of this project relies on extracting concepts to describe
a text, having a human description (i.e. a gloss or a
  \emph{headword}) for each concept is an obligation. However, most
of the experiments proved that there is a majority of \emph{anonymous}
concepts. Even if they remain important in the construction of the
conceptual hierarchy, they must not be displayed to the end user.

\subsection*{Multiple Roots and Orphan Nodes}
\label{sec:edr:mult-roots-orph}

The EDR documentation is not really clear on what are the
\gloss[word]{ontology}'s roots. A note in the example about the basic
words states that all concepts are located below the concept
\emph{\#3aa966}.  This does not seem to be strictly true. Indeed, all
concepts dominated by this root have a \textbf{conceptType} of
``SIMPLE'' but during our experimentation, a second root, the
concept \emph{\#2f3526}, has been identified.  All the concepts
dominated by this root have a \textbf{conceptType} of ``TECHNICAL''
which might explains the distinction between the two separate
hierarchy.

Counting all the concepts located under these two roots gave a sum of
474'169 concepts, whereas simply counting the number of entries in the
\emph{headconcept} databases gave 488'732. 14'563 concepts are neither
under the concept \emph{\#3aa966} nor under the concept
\emph{\#2f3526}. These orphan concepts are not linked in the database
to any other concepts.

\subsection*{non-summable frequencies}
\label{sec:edr:non-summ-freq}

Each word in the EDR English dictionary has information on its
frequency in the EDR corpus. However, a few issues arise; let's take
 table \ref{sec:edr:alabama} as an example.

\begin{table}[htb]
\centering
\begin{tabular}{|c|c|c|c|c|}
\hline
\textbf{invariable portion} & \textbf{POS} & \textbf{concept} &
\textbf{$F_{word}^{concept}$} & \textbf{$F_{word}$} \\
\hline
Alabama & Common Noun & 581 & 2 & 5 \\
Alabama & Common Noun & 582 & 0 & 4 \\
Alabama & Common Noun & 582 & 0 & 5 \\
Alabama & Common Noun & 583 & 0 & 5 \\
Alabama & Proper Noun & 362294 & 10 & 10 \\
\hline
\end{tabular}
\caption{The Alabama entries in the EDR database}
\label{sec:edr:alabama}
\end{table}

The word "Alabama" has five entries corresponding to five different
concepts.  EDR gives a total Frequency for the word ($F_{word}$)
independent from other information (i.e. how many times this
  word appears in the corpus) and a Frequency ($F_{word}^{concept}$)
for the word with one particular concept (i.e. how many times the word
  was encountered having this concept). Ideally, the
$F_{word}^{concept}$ should sum up to $F_{word}$. However, a few values are
wrong (speaking of table \ref{sec:edr:alabama}):
\begin{itemize}
\item most of the $F_{word}^{concept}$ are null. This means that the
  word with this concept has never been found in the corpus, but exists
  anyway in the database. All frequencies should at least
  be one in order to be used as a realistic value of what is
  encountered in usual texts.
\item $F_{word}$ is not always the same (5, 4 and 10). A word should
  have a constant frequency that does not depend on other information
  than the lexicographic form of the \gloss[word]{lemma}.
\item the $F_{word}^{concept}$ do not sum to any $F_{word}$.
\end{itemize}

\section{Interesting Tables}
\label{sec:edr:interesting-tables}

The EDR database is divided amongst a few tables, the most important of them
representing one of the original \emph{dictionary} structures present
in the textual files, the others bearing textual constants.
For example, a \gloss[word]{lemma} entry would be searched in the
\textbf{HeadWords} table, that would return an index to access the
English words table (\textbf{EWD}).

\subsection*{EWD}
\label{sec:edr:ewd}

The \textbf{EWD} table represents the English words dictionary and
contains the basic grammatical information and the indexes of the
attached concepts. This table is mostly used to generate the lexicon
(see section \ref{sec:ext:token}) and to create the initial bag of
concepts (see section \ref{sec:chap:docum-repr}). The \gloss[word]{lemma} and the
\gloss[short]{pos} are respectively searched in the \textbf{HeadWords}
and \textbf{PartOfSpeechs} tables, the indexes of which are used to select the
right row in \textbf{EWD}.

\subsection*{CPH}
\label{sec:edr:cph}

The \textbf{CPH} table is one of the concept dictionaries; it
basically describes each concept with an \emph{explanation} and a
\emph{headconcept}. The \emph{id} field of this table is used in most
representations as it makes the link between the \textbf{EWD} table,
the \textbf{CPC} table and this one.

\subsection*{CPC,CPT}
\label{sec:edr:cpc}

The \textbf{CPC} table provides the
\gloss[word]{hyponym}/\gloss[word]{hypernym} link. Each entry has an
\emph{id}  -- relating to the relation, nothing to do with
  \textbf{CPH} ids-- , one \emph{subconcept} id and one
\emph{superconcept} id. The entry must be interpreted as ``subconcept
$\leftarrow$ superconcept'' and the \emph{id} column ignored.

The \textbf{CPT} table represents the other possible links between
words and has not been used in this project.

\section{New Tables}
\label{sec:new-tables}

For faster computation in the extraction algorithm (see section
\ref{cha:extraction}), offline computations were performed on the
existing tables. Results from these computations are stored in two
new tables. Here is a quick description of the new data introduced.

\subsection*{the cptscore table}
\label{sec:cptscore}

Because we are in a \gloss[short]{DAG} there is more than one path
between two concepts. Especially, between a concept and the root, and
between a concept and its covered leaves.

Multiple inheritance in an ontology is not really intuitive. We believe that
if a same concept have two paths to the root, then the meaning of this
concept on a path is not the same as the one on the other path: the
entry is unique in the EDR hierarchy but represents two different
concepts.

Therefore, the number of leaves covered by a concept must take into
account all the {meanings} of a leaf entry. If a concept has one leaf
but three path to this leaf, it is considered to cover three different
leaves.

In the same way, we cannot directly compute the path length from one
concept to another. Therefore -- because we don't know which meaning
an entry is considered for -- we have to compute an average distance.
Moreover, we are interested in the normalized distance between two
concepts. In an ontology, there is often some parts that are more
informed -- with a better granularity in the concept hierarchy -- than
others because the creators have more knowledge on these parts. A
normalized distance between concepts is then more robust to these
variations of granularity. 

\begin{figure}[htbp]
  \centering \includegraphics[width=.5\textwidth]{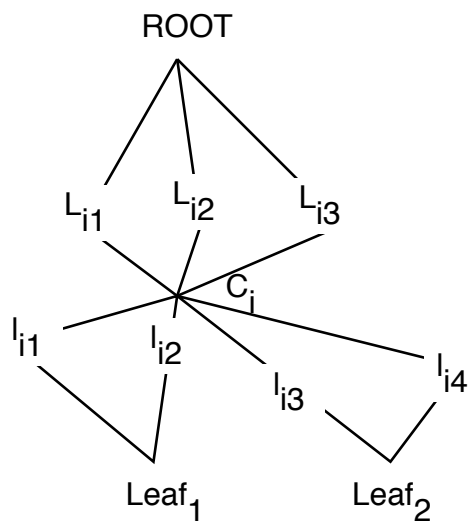}
  \caption{Path from a concept $c_i$}
  \label{fig:edr:path}
\end{figure}

We are interested by the normalized average distance in the
\gloss[short]{DAG} between each concept and the root.  We can directly
compute from the hierarchy (see figure \ref{fig:edr:path}):
\begin{itemize}
\item[$L_{i,j}$] the distance from the concept $c_i$ to the root
  following the path $j$ ,
\item[$l_{i,k}$] the distance from a concept $c_i$ to one of its leaves
  with the path $k$,
\item $N$ the total number of path to the root,
\item $n$ the total number of path to the leaves,
\end{itemize}

We can define the normalized distance to the root as:
$$
D(c_i) = \frac{1}{N \times n} \times \sum_{j=0}^{N}\sum_{k=0}^{n} (\frac{L_{ij}}{L_{ij}+l_{ik}})
$$

In the same way, we can define the average distance to the covered
leaves as:
$$
d(c_i) = \frac{1}{N \times n} \times \sum_{j=0}^{N}\sum_{k=0}^{n} (\frac{l_{ij}}{L_{ij}+l_{ik}})
$$

\subsection*{the concept2leaf table}
\label{sec:cpt2leaf}

Another important value that is needed for the topic extraction
algorithm (see section \ref{sec:concept-scoring}) is the distance
between a concept and each of its leaves. Unlike the distance
discussed above, the raw edge count between a concept and each of its
covered list must be stored.

Because we can have a lot of path from a concept to all its leaves,
this new table is huge. For example, the root has \textbf{376'434}
entry in this table.

\section{Corpus and Lexicon}
\label{sec:corpus-lexicon}

\subsection*{Lexicon Creation}
\label{sec:lexicon-crea}

The language processing toolkit developed at LIA \cite{SLPTK} provides
the adequate structures for the creation of a specialized tokenizer
and lemmatizer. The first step of this project was therefore dedicated
to the production of a \gloss[word]{lexicon} that could be used by
this tool; the lexicon had to contain each possible words -- in its
inflected form -- and the corresponding \gloss[none]{lemma}lemmas. The
EDR database does not contain the inflected forms of words, but only
\gloss[none]{lemma}lemmas with inflection patterns. Therefore, the
\gloss[word]{lexicon} was derived from this information.

The EDR dictionary contains 52 possible inflection patterns. Each
entry in EDR -- the lemmas -- is assigned a pattern in function of its
\gloss[short]{pos}, however some lemmas have irregular flexions (e.g.
``is'' cannot be inflected according to regular patterns). In this
case, each possible flexion of the lemma is introduced in the database
(e.g. ``is'', ``be'', ``was'' and ``been'' will be the three entries
for the verb ``is'').  With 52 inflection patterns applied on the
122'498 regular entries, plus the 132'982 irregular forms already
present in the EDR dictionary, 394'703 different flexions are
generated for the lexicon. Which is approximately 2 flexions by
regular entry.  For example ``call'' has the following inflection
patterns:
\begin{itemize}
\item "s-ed" "-" "s" "ed" "ed" "ing" as a Verb
\item "s" "-" "s" as a Noun
\end{itemize}
and will be inflected:
\begin{itemize}
\item called (Verbs)
\item calling (Verbs)
\item calls (Nouns)
\item calls (Verbs)
\item call (Nouns)
\item call (Verbs)
\end{itemize}

\subsection*{Corpus Transformation}
\label{sec:corp-transf}

The EDR dictionary comes with a corpus of 125'814 short sentences extracted from
different journals. These sentences are annotated with different
information such as the sentence segmentation, concepts attached to
each words or -- which is more interesting in this project -- the part of
speech category of each word.

\begin{verbatim}
ECO0000021  0070000034e6    The Independent 0.8 per cent and now 0.5
per cent.  

{   1   0.8 0.8 NUM "=N 0.8"    2   _   _   BLNK    2dc2ed  3
   per_cent    per_cent    NOUN    "=Z percent"    4   _   _   BLNK
   2dc2ed  5   and and CONJ    2dc2f0  6   _   _   BLNK    2dc2ed  7
   now now ADV 0cb281  8   _   _   BLNK    2dc2ed  9   0.5 0.5 NUM "=N
   0.5"    10  _   _   BLNK    ""  11  per_cent    per_cent    NOUN
   "=Z percent"    12  .   .   PUNC    2dc2e5 
}

   /1:0.8/2:_/3:per_cent/4:_/5:and/6:_/7:now/8:_/9:0.5/10:_/11:per_cent/12:./ 

(S(t(S(S(t(M(S(t(W 1 "0.8"))(W 2 "_"))(t(W 3 "per_cent"))))(W 4
"_"))(t(S(S(t(W 5 "and"))(W 6 "_"))(t(M(S(t(W 7 "now"))(W 8
"_"))(t(M(S(t(W 9 "0.5"))(W 10 "_"))(t(W 11 "per_cent"))))))))))(W 12
"."))

 [[main 11:"per_cent":"=Z percent"][time 7:"now":0cb281][number
 9:"0.5":"=N 0.5"][and [[main 3:"per_cent":"=Z percent"][number
 1:"0.8":"=N 0.8"]]]] DATE="95/7/12"
\end{verbatim}

In the perspective of using this annotation as a training base for a
Part of Speech tagger \cite{treetagger}, the corpus was transformed to
the correct input format.

However, the set of \gloss[short]{pos} tag in the EDR corpus is quite
limited: noun, pronoun, demonstrative, word of negation, question
word, intransitive verb, transitive verb, verb, Be-verb,
adjective, adverb, adverbial particle, conjunction, prefix,
suffix, article, auxiliary, verb, preposition, interjection,
blank, space, punctuation symbol, symbol unit, number.

To get a more robust tagger, this tag set must be extended to retain
more information on the grammatical context. For example, it is
important to know if a verb is at a passive or present form.
Therefore, the following set of tags have been chosen:
\begin{itemize}
\item an article (ART)
\item an auxiliary (AUX)
\item a comparative adjective (Adj\_Comp)
\item a positive adjective (Adj\_Pos)
\item a superlative adjective (Adj\_Super)
\item a comparative adverb (Adv\_Comp)
\item a positive adverb (Adv\_Pos)
\item a superlative adverb (Adv\_Super)
\item the verb be (BE)
\item a conjunction (CONJ)
\item a demonstrative (DEMO)
\item Indefinite Pronoun (INDEF)
\item interjection (ITJ)
\item plural noun (NP)
\item singular noun (NS)
\item number (NUM)
\item prefix (PF)
\item proper noun (PN)
\item preposition (PREP)
\item pronoun (PRON)
\item particle (PTCL)
\item punctuation (PUNC)
\item symbol (SYM)
\item to (TO)
\item unknown (UKN)
\item unit (UNIT)
\item basic verb form (VB)
\item verb, 3rd person singular (V\_3RD)
\item verb, past form (V\_PF)
\item verb, past participle (V\_PaP)
\item verb, present participle (V\_PrP)
\item verb, present form (V\_Pres)
\item wh. pronoun (WH)
\end{itemize}

Transformation from the original set of tags have been made by
following simple heuristics based on the word suffixes -- e.g a noun
followed by ``s'' is a plural noun. When these rules cannot resolve
the ambiguity on a tag translation -- for example, a verb that has the suffix ``ed'' can
either be a past form or a past participle -- an existing tagger is used to tag
the ambiguous word.

\chapter{Segmentation}
\label{cha:segmentation}

As seen in the section \ref{sec:state-art:segmentation}, two major
segmentation techniques have proved to be efficient. In this project,
where the extraction process is bound to linguistic resources, keeping
the segmentation \emph{language independent} is not a priority. The
\gloss[word]{lexicalchains} technique or an extended \gloss[word]{bow}
representation might therefore be good bases to use the available
\gloss[word]{ontology}. However, the aim of this project is more to
develop a novel topic extraction technique, thus it was decided to
implement a simple Text Tiling method as proposed by M.Hearst
\cite{hearst94multiparagraph} to avoid spending time on the issues
raised by the use of the EDR \gloss[word]{ontology} (see chapter
\ref{cha:edr-dict-ontol} and section \ref{sec:ext:preprocessing}). The
method used is an adaptation of M.Hearst original method to embody the
results and recommendations made in \cite{ferret98thematic} (for
equations (\ref{eq:seg:weight}) and (\ref{eq:dice})) and
\cite{richmond97detecting} (for the smoothing, section
\ref{sec:smooth}).

\begin{itemize}
\item The original text is first tokenized and common words (coming
  from a \emph{stop-list}) are removed. No other preprocessing is
  performed.
\item The text is then divided in preliminary windows, each
  represented by a weighted \gloss[word]{bow}.
\item The similarity between each adjacent window is computed and
  boundaries are placed in accordance to these values.
\end{itemize}

This process is supported by a linear discourse model inspired by
Chafe's notion of Flow Model \cite{chafe79flow}. This model suggests
that the author, when moving from topic to topic, induces a change in
the thematic components (i.e. topical words, subject, etc\ldots). The
interesting point is that such change is accompanied by a decrease in
the lexical cohesion of the text. The Text Tiling algorithm tries to
determine this change by using the term repetitions as an indicator of
this lexical cohesion. The similarity computation between text windows
is then a computation of the lexical cohesion between consecutive
blocks of the document. When this similarity reaches a local minimum,
a topic shift is detected.

\section{Preprocessing}
\label{sec:seg:preprocessing}

Because the segmentation is based on geometric vector distance, it
could be interesting to have the highest vector density possible. Indeed, the
distance between each vector is a representation of the similitude of
their textual content, therefore if the vectors are dense, it will be
easier to determine an accurate distance. To obtain denser
vectors, it is often better to index them by the word
\gloss[word]{lemma} than by the inflected form (i.e. running is
similar in meaning to run).

However, to find the correct \gloss[word]{lemma} for a word, it is
important to know its \gloss[long]{pos} category and the possible
inflections for each category. In the EDR database, the inflection
patterns are available and it is easy to derive a
\gloss[word]{lexicon} from them; however, at the time the experiments were
made on segmentation, no \gloss[short]{pos} tagger for the EDR lexicon
was available.

Therefore, having the \gloss[none]{}lemmas without the possibility of
disambiguation given by a tagger is not really helpful. In fact,
without disambiguation, a considerable irrelevant noise -- more than
  one \gloss[word]{lemma} per word -- will be inserted in the vectors,
increasing the computation inaccuracy. Moreover, M.Hearst has
demonstrated that the use of a simple \gloss[word]{stoplist} was
enough to remove the common words and to give good results.

\section{Vectorization And Weighting}
\label{sec:seg:vector}

After the preprocessing, the document's vocabulary is computed, which
 contains all the possible words in the document and
provides a basis for the vector representation. The document is then
segmented into $N$ windows containing the same number of words (a parameter of
the algorithm). Each of these windows is represented by a vector $G_i
= (g_{i1}, g_{i2}, \ldots , g_{in})$ where $g_{ij}$ is the number of
occurrences in the window $i$ of the $j^{st}$ token in the document's
vocabulary and $n$ the size of this vocabulary.

Each element of the vector is then weighted using the
\gloss[word]{TFIDF} method.  The weight $w_{ij}$ is a ratio between
the inner-window frequency and the inner-document frequency:
\begin{equation}
  \label{eq:seg:weight}
  w_{ij} = g_{ij} \times \log(\frac{N}{df_j})
\end{equation}

$g_{ij}$ being the number of occurrences of the word $j$ in the
$i^{st}$ window, $N$ the total number of windows and $df_j$ the number
of windows where the word $j$ appears. Hence, the weight of a word
spread along the whole document will be lower than the weight of one
that occurs in a few windows.

\section{Similarity Metric}
\label{sec:seg:measure}

Once the vectors have been weighted, a curve is plotted with the
similarity between each adjacent windows along the window number axis.

The similarity is computed using the \emph{Dice Coefficient} defined
by:
\begin{equation}
  \label{eq:dice}
  D(G_1,G_2) = \frac{2 \times \sum_{i=1}^{n} (w_{1i} \times w_{2i})}{\sum_{i=1}^{n} w_{1i}^2 \times \sum_{i=1}^{n} w_{2i}^2}
\end{equation}

\section{Smoothing}
\label{sec:smooth}

\begin{figure}[htbp]
  \centering \setlength{\unitlength}{0.240900pt}
\ifx\plotpoint\undefined\newsavebox{\plotpoint}\fi
\begin{picture}(1500,900)(0,0)
\sbox{\plotpoint}{\rule[-0.200pt]{0.400pt}{0.400pt}}%
\put(60.0,82.0){\rule[-0.200pt]{4.818pt}{0.400pt}}
\put(1419.0,82.0){\rule[-0.200pt]{4.818pt}{0.400pt}}
\put(60.0,160.0){\rule[-0.200pt]{4.818pt}{0.400pt}}
\put(1419.0,160.0){\rule[-0.200pt]{4.818pt}{0.400pt}}
\put(60.0,238.0){\rule[-0.200pt]{4.818pt}{0.400pt}}
\put(1419.0,238.0){\rule[-0.200pt]{4.818pt}{0.400pt}}
\put(60.0,315.0){\rule[-0.200pt]{4.818pt}{0.400pt}}
\put(1419.0,315.0){\rule[-0.200pt]{4.818pt}{0.400pt}}
\put(60.0,393.0){\rule[-0.200pt]{4.818pt}{0.400pt}}
\put(1419.0,393.0){\rule[-0.200pt]{4.818pt}{0.400pt}}
\put(60.0,471.0){\rule[-0.200pt]{4.818pt}{0.400pt}}
\put(1419.0,471.0){\rule[-0.200pt]{4.818pt}{0.400pt}}
\put(60.0,549.0){\rule[-0.200pt]{4.818pt}{0.400pt}}
\put(1419.0,549.0){\rule[-0.200pt]{4.818pt}{0.400pt}}
\put(60.0,627.0){\rule[-0.200pt]{4.818pt}{0.400pt}}
\put(1419.0,627.0){\rule[-0.200pt]{4.818pt}{0.400pt}}
\put(60.0,704.0){\rule[-0.200pt]{4.818pt}{0.400pt}}
\put(1419.0,704.0){\rule[-0.200pt]{4.818pt}{0.400pt}}
\put(60.0,782.0){\rule[-0.200pt]{4.818pt}{0.400pt}}
\put(1419.0,782.0){\rule[-0.200pt]{4.818pt}{0.400pt}}
\put(60.0,860.0){\rule[-0.200pt]{4.818pt}{0.400pt}}
\put(1419.0,860.0){\rule[-0.200pt]{4.818pt}{0.400pt}}
\put(60.0,82.0){\rule[-0.200pt]{0.400pt}{4.818pt}}
\put(60,41){\makebox(0,0){ 0}}
\put(60.0,840.0){\rule[-0.200pt]{0.400pt}{4.818pt}}
\put(306.0,82.0){\rule[-0.200pt]{0.400pt}{4.818pt}}
\put(306,41){\makebox(0,0){ 5}}
\put(306.0,840.0){\rule[-0.200pt]{0.400pt}{4.818pt}}
\put(553.0,82.0){\rule[-0.200pt]{0.400pt}{4.818pt}}
\put(553,41){\makebox(0,0){ 10}}
\put(553.0,840.0){\rule[-0.200pt]{0.400pt}{4.818pt}}
\put(799.0,82.0){\rule[-0.200pt]{0.400pt}{4.818pt}}
\put(799,41){\makebox(0,0){ 15}}
\put(799.0,840.0){\rule[-0.200pt]{0.400pt}{4.818pt}}
\put(1045.0,82.0){\rule[-0.200pt]{0.400pt}{4.818pt}}
\put(1045,41){\makebox(0,0){ 20}}
\put(1045.0,840.0){\rule[-0.200pt]{0.400pt}{4.818pt}}
\put(1291.0,82.0){\rule[-0.200pt]{0.400pt}{4.818pt}}
\put(1291,41){\makebox(0,0){ 25}}
\put(1291.0,840.0){\rule[-0.200pt]{0.400pt}{4.818pt}}
\put(60.0,82.0){\rule[-0.200pt]{332.201pt}{0.400pt}}
\put(1439.0,82.0){\rule[-0.200pt]{0.400pt}{187.420pt}}
\put(60.0,860.0){\rule[-0.200pt]{332.201pt}{0.400pt}}
\put(60.0,82.0){\rule[-0.200pt]{0.400pt}{187.420pt}}
\multiput(60.58,82.00)(0.498,2.380){95}{\rule{0.120pt}{1.994pt}}
\multiput(59.17,82.00)(49.000,227.862){2}{\rule{0.400pt}{0.997pt}}
\multiput(109.00,312.92)(1.145,-0.496){41}{\rule{1.009pt}{0.120pt}}
\multiput(109.00,313.17)(47.906,-22.000){2}{\rule{0.505pt}{0.400pt}}
\multiput(159.58,292.00)(0.498,4.189){95}{\rule{0.120pt}{3.431pt}}
\multiput(158.17,292.00)(49.000,400.880){2}{\rule{0.400pt}{1.715pt}}
\multiput(208.58,682.64)(0.498,-5.134){95}{\rule{0.120pt}{4.182pt}}
\multiput(207.17,691.32)(49.000,-491.321){2}{\rule{0.400pt}{2.091pt}}
\multiput(257.58,196.98)(0.498,-0.787){95}{\rule{0.120pt}{0.729pt}}
\multiput(256.17,198.49)(49.000,-75.488){2}{\rule{0.400pt}{0.364pt}}
\multiput(306.00,121.92)(0.610,-0.498){79}{\rule{0.588pt}{0.120pt}}
\multiput(306.00,122.17)(48.780,-41.000){2}{\rule{0.294pt}{0.400pt}}
\multiput(356.58,82.00)(0.498,1.733){95}{\rule{0.120pt}{1.480pt}}
\multiput(355.17,82.00)(49.000,165.929){2}{\rule{0.400pt}{0.740pt}}
\multiput(405.58,251.00)(0.498,1.157){95}{\rule{0.120pt}{1.022pt}}
\multiput(404.17,251.00)(49.000,110.878){2}{\rule{0.400pt}{0.511pt}}
\multiput(454.58,361.38)(0.498,-0.664){95}{\rule{0.120pt}{0.631pt}}
\multiput(453.17,362.69)(49.000,-63.691){2}{\rule{0.400pt}{0.315pt}}
\multiput(503.58,299.00)(0.498,2.554){97}{\rule{0.120pt}{2.132pt}}
\multiput(502.17,299.00)(50.000,249.575){2}{\rule{0.400pt}{1.066pt}}
\multiput(553.58,550.76)(0.498,-0.551){95}{\rule{0.120pt}{0.541pt}}
\multiput(552.17,551.88)(49.000,-52.878){2}{\rule{0.400pt}{0.270pt}}
\multiput(602.58,491.10)(0.498,-2.267){95}{\rule{0.120pt}{1.904pt}}
\multiput(601.17,495.05)(49.000,-217.048){2}{\rule{0.400pt}{0.952pt}}
\multiput(651.58,271.59)(0.498,-1.815){95}{\rule{0.120pt}{1.545pt}}
\multiput(650.17,274.79)(49.000,-173.793){2}{\rule{0.400pt}{0.772pt}}
\multiput(700.58,101.00)(0.498,2.252){97}{\rule{0.120pt}{1.892pt}}
\multiput(699.17,101.00)(50.000,220.073){2}{\rule{0.400pt}{0.946pt}}
\multiput(750.58,321.26)(0.498,-1.003){95}{\rule{0.120pt}{0.900pt}}
\multiput(749.17,323.13)(49.000,-96.132){2}{\rule{0.400pt}{0.450pt}}
\multiput(799.58,227.00)(0.498,4.826){95}{\rule{0.120pt}{3.937pt}}
\multiput(798.17,227.00)(49.000,461.829){2}{\rule{0.400pt}{1.968pt}}
\multiput(848.58,685.40)(0.498,-3.387){95}{\rule{0.120pt}{2.794pt}}
\multiput(847.17,691.20)(49.000,-324.201){2}{\rule{0.400pt}{1.397pt}}
\multiput(897.58,362.73)(0.498,-1.164){97}{\rule{0.120pt}{1.028pt}}
\multiput(896.17,364.87)(50.000,-113.866){2}{\rule{0.400pt}{0.514pt}}
\multiput(947.58,244.86)(0.498,-1.733){95}{\rule{0.120pt}{1.480pt}}
\multiput(946.17,247.93)(49.000,-165.929){2}{\rule{0.400pt}{0.740pt}}
\multiput(996.58,82.00)(0.498,2.339){95}{\rule{0.120pt}{1.961pt}}
\multiput(995.17,82.00)(49.000,223.929){2}{\rule{0.400pt}{0.981pt}}
\multiput(1045.58,307.04)(0.498,-0.766){95}{\rule{0.120pt}{0.712pt}}
\multiput(1044.17,308.52)(49.000,-73.522){2}{\rule{0.400pt}{0.356pt}}
\multiput(1094.58,235.00)(0.498,1.949){97}{\rule{0.120pt}{1.652pt}}
\multiput(1093.17,235.00)(50.000,190.571){2}{\rule{0.400pt}{0.826pt}}
\multiput(1144.58,416.83)(0.498,-3.562){95}{\rule{0.120pt}{2.933pt}}
\multiput(1143.17,422.91)(49.000,-340.913){2}{\rule{0.400pt}{1.466pt}}
\multiput(1193.58,82.00)(0.498,2.318){95}{\rule{0.120pt}{1.945pt}}
\multiput(1192.17,82.00)(49.000,221.963){2}{\rule{0.400pt}{0.972pt}}
\multiput(1242.00,306.93)(3.696,-0.485){11}{\rule{2.900pt}{0.117pt}}
\multiput(1242.00,307.17)(42.981,-7.000){2}{\rule{1.450pt}{0.400pt}}
\multiput(1291.58,301.00)(0.498,4.900){97}{\rule{0.120pt}{3.996pt}}
\multiput(1290.17,301.00)(50.000,478.706){2}{\rule{0.400pt}{1.998pt}}
\multiput(1341.58,777.18)(0.498,-3.151){95}{\rule{0.120pt}{2.606pt}}
\multiput(1340.17,782.59)(49.000,-301.591){2}{\rule{0.400pt}{1.303pt}}
\multiput(1390.58,481.00)(0.498,0.879){95}{\rule{0.120pt}{0.802pt}}
\multiput(1389.17,481.00)(49.000,84.335){2}{\rule{0.400pt}{0.401pt}}
\put(60,82){\circle{12}}
\put(109,314){\circle{12}}
\put(159,292){\circle{12}}
\put(208,700){\circle{12}}
\put(257,200){\circle{12}}
\put(306,123){\circle{12}}
\put(356,82){\circle{12}}
\put(405,251){\circle{12}}
\put(454,364){\circle{12}}
\put(503,299){\circle{12}}
\put(553,553){\circle{12}}
\put(602,499){\circle{12}}
\put(651,278){\circle{12}}
\put(700,101){\circle{12}}
\put(750,325){\circle{12}}
\put(799,227){\circle{12}}
\put(848,697){\circle{12}}
\put(897,367){\circle{12}}
\put(947,251){\circle{12}}
\put(996,82){\circle{12}}
\put(1045,310){\circle{12}}
\put(1094,235){\circle{12}}
\put(1144,429){\circle{12}}
\put(1193,82){\circle{12}}
\put(1242,308){\circle{12}}
\put(1291,301){\circle{12}}
\put(1341,788){\circle{12}}
\put(1390,481){\circle{12}}
\put(1439,567){\circle{12}}
\put(330,820){\circle{12}}
\put(60.0,82.0){\rule[-0.200pt]{332.201pt}{0.400pt}}
\put(1439.0,82.0){\rule[-0.200pt]{0.400pt}{187.420pt}}
\put(60.0,860.0){\rule[-0.200pt]{332.201pt}{0.400pt}}
\put(60.0,82.0){\rule[-0.200pt]{0.400pt}{187.420pt}}
\put(750,0){\makebox(0,0){segment index}}%
\put(20,500){%
\special{ps: gsave currentpoint currentpoint translate
270 rotate neg exch neg exch translate}%
\makebox(0,0)[b]{\shortstack{similarity}}%
\special{ps: currentpoint grestore moveto}%
}%
\end{picture}
  \caption[Similarity between consecutive windows.]{Similarity between consecutive windows. (window size: 25,
    average real segment length: 240 words)}
  \label{fig:seg:nonsmooth}
\end{figure}
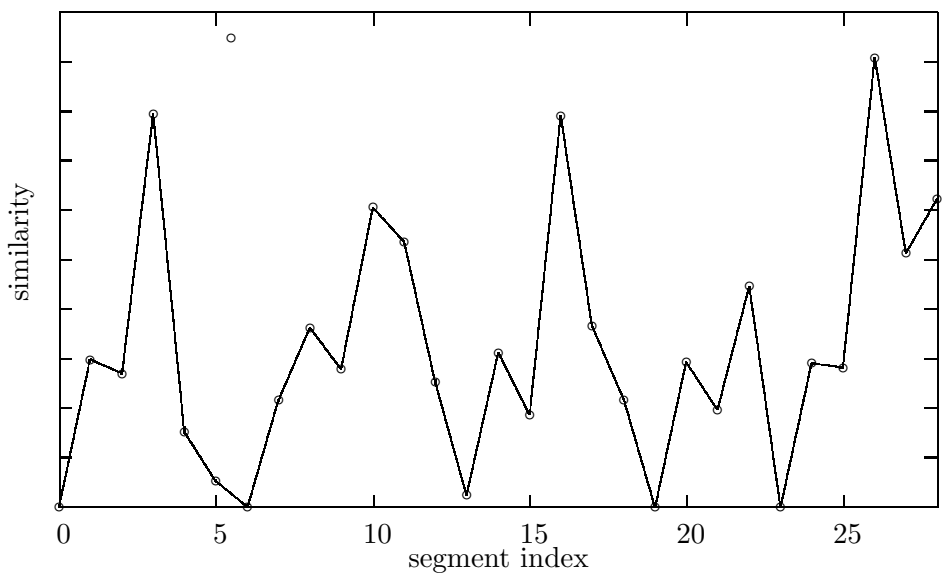

A low similarity between two windows can be interpreted as a thematic
change whereas a high value represents a continuity in the windows'
topic. The figure \ref{fig:seg:nonsmooth} is an example of the
similarity variation encountered.

The boundary detection (see next section) is based on the local minima
but, as illustrated in figure \ref{fig:seg:nonsmooth}, the curve is
quite chaotic and contains irrelevant minima. For example, even if the last
point has a hight similarity value, it is a local minimum. To
prevent the detection of irrelevant boundaries, a simple smoothing is
performed on the curve to isolate  high value changes from 
local perturbations.

For each point $P_i$ of the graph, the bisection B of the points
$P_{i-1}$ and $P_{i+1}$ is computed and the point $P_i$ is moved in the
direction of B by a constant amount. This smoothing is performed on
each point in parallel and repeated several times.

\begin{figure}[htb]
\begin{center}
  \includegraphics[width=.5\textwidth]{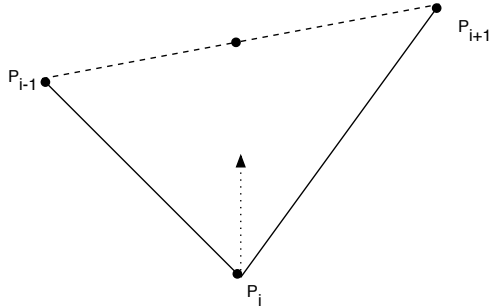}
\caption{Smoothing}
\label{fig:seg:smooth}
\end{center}
\end{figure}

Another solution to smooth the similarity curve would have been to
overlap each consecutive window so the end of one would have been the
start of the next one. Even if this change has not been tested, it is
believed to be more robust as the constant shift imposed in the
previously described smoothing is another non obvious parameter to the
algorithm.

\section{Boundary Detection}
\label{sec:seg:bound}

\begin{figure}[htbp]
  \centering \setlength{\unitlength}{0.240900pt}
\ifx\plotpoint\undefined\newsavebox{\plotpoint}\fi
\begin{picture}(1500,900)(0,0)
\sbox{\plotpoint}{\rule[-0.200pt]{0.400pt}{0.400pt}}%
\put(60.0,82.0){\rule[-0.200pt]{4.818pt}{0.400pt}}
\put(1419.0,82.0){\rule[-0.200pt]{4.818pt}{0.400pt}}
\put(60.0,179.0){\rule[-0.200pt]{4.818pt}{0.400pt}}
\put(1419.0,179.0){\rule[-0.200pt]{4.818pt}{0.400pt}}
\put(60.0,277.0){\rule[-0.200pt]{4.818pt}{0.400pt}}
\put(1419.0,277.0){\rule[-0.200pt]{4.818pt}{0.400pt}}
\put(60.0,374.0){\rule[-0.200pt]{4.818pt}{0.400pt}}
\put(1419.0,374.0){\rule[-0.200pt]{4.818pt}{0.400pt}}
\put(60.0,471.0){\rule[-0.200pt]{4.818pt}{0.400pt}}
\put(1419.0,471.0){\rule[-0.200pt]{4.818pt}{0.400pt}}
\put(60.0,568.0){\rule[-0.200pt]{4.818pt}{0.400pt}}
\put(1419.0,568.0){\rule[-0.200pt]{4.818pt}{0.400pt}}
\put(60.0,666.0){\rule[-0.200pt]{4.818pt}{0.400pt}}
\put(1419.0,666.0){\rule[-0.200pt]{4.818pt}{0.400pt}}
\put(60.0,763.0){\rule[-0.200pt]{4.818pt}{0.400pt}}
\put(1419.0,763.0){\rule[-0.200pt]{4.818pt}{0.400pt}}
\put(60.0,860.0){\rule[-0.200pt]{4.818pt}{0.400pt}}
\put(1419.0,860.0){\rule[-0.200pt]{4.818pt}{0.400pt}}
\put(60.0,82.0){\rule[-0.200pt]{0.400pt}{4.818pt}}
\put(60,41){\makebox(0,0){ 0}}
\put(60.0,840.0){\rule[-0.200pt]{0.400pt}{4.818pt}}
\put(306.0,82.0){\rule[-0.200pt]{0.400pt}{4.818pt}}
\put(306,41){\makebox(0,0){ 5}}
\put(306.0,840.0){\rule[-0.200pt]{0.400pt}{4.818pt}}
\put(553.0,82.0){\rule[-0.200pt]{0.400pt}{4.818pt}}
\put(553,41){\makebox(0,0){ 10}}
\put(553.0,840.0){\rule[-0.200pt]{0.400pt}{4.818pt}}
\put(799.0,82.0){\rule[-0.200pt]{0.400pt}{4.818pt}}
\put(799,41){\makebox(0,0){ 15}}
\put(799.0,840.0){\rule[-0.200pt]{0.400pt}{4.818pt}}
\put(1045.0,82.0){\rule[-0.200pt]{0.400pt}{4.818pt}}
\put(1045,41){\makebox(0,0){ 20}}
\put(1045.0,840.0){\rule[-0.200pt]{0.400pt}{4.818pt}}
\put(1291.0,82.0){\rule[-0.200pt]{0.400pt}{4.818pt}}
\put(1291,41){\makebox(0,0){ 25}}
\put(1291.0,840.0){\rule[-0.200pt]{0.400pt}{4.818pt}}
\put(60.0,82.0){\rule[-0.200pt]{332.201pt}{0.400pt}}
\put(1439.0,82.0){\rule[-0.200pt]{0.400pt}{187.420pt}}
\put(60.0,860.0){\rule[-0.200pt]{332.201pt}{0.400pt}}
\put(60.0,82.0){\rule[-0.200pt]{0.400pt}{187.420pt}}
\put(80,820){\makebox(0,0)[l]{smoothed}}
\put(260.0,820.0){\rule[-0.200pt]{24.090pt}{0.400pt}}
\put(60,276){\usebox{\plotpoint}}
\multiput(60.58,276.00)(0.498,0.982){95}{\rule{0.120pt}{0.884pt}}
\multiput(59.17,276.00)(49.000,94.166){2}{\rule{0.400pt}{0.442pt}}
\multiput(109.58,372.00)(0.498,1.677){97}{\rule{0.120pt}{1.436pt}}
\multiput(108.17,372.00)(50.000,164.020){2}{\rule{0.400pt}{0.718pt}}
\multiput(159.58,539.00)(0.498,1.229){95}{\rule{0.120pt}{1.080pt}}
\multiput(158.17,539.00)(49.000,117.759){2}{\rule{0.400pt}{0.540pt}}
\multiput(208.58,650.62)(0.498,-2.411){95}{\rule{0.120pt}{2.018pt}}
\multiput(207.17,654.81)(49.000,-230.811){2}{\rule{0.400pt}{1.009pt}}
\multiput(257.58,420.30)(0.498,-0.993){95}{\rule{0.120pt}{0.892pt}}
\multiput(256.17,422.15)(49.000,-95.149){2}{\rule{0.400pt}{0.446pt}}
\multiput(306.58,324.89)(0.498,-0.509){97}{\rule{0.120pt}{0.508pt}}
\multiput(305.17,325.95)(50.000,-49.946){2}{\rule{0.400pt}{0.254pt}}
\multiput(356.00,276.58)(1.377,0.495){33}{\rule{1.189pt}{0.119pt}}
\multiput(356.00,275.17)(46.532,18.000){2}{\rule{0.594pt}{0.400pt}}
\multiput(405.00,294.58)(0.556,0.498){85}{\rule{0.545pt}{0.120pt}}
\multiput(405.00,293.17)(47.868,44.000){2}{\rule{0.273pt}{0.400pt}}
\multiput(454.58,338.00)(0.498,1.157){95}{\rule{0.120pt}{1.022pt}}
\multiput(453.17,338.00)(49.000,110.878){2}{\rule{0.400pt}{0.511pt}}
\multiput(503.00,451.58)(0.966,0.497){49}{\rule{0.869pt}{0.120pt}}
\multiput(503.00,450.17)(48.196,26.000){2}{\rule{0.435pt}{0.400pt}}
\multiput(553.58,474.28)(0.498,-0.694){95}{\rule{0.120pt}{0.655pt}}
\multiput(552.17,475.64)(49.000,-66.640){2}{\rule{0.400pt}{0.328pt}}
\multiput(602.58,405.81)(0.498,-0.838){95}{\rule{0.120pt}{0.769pt}}
\multiput(601.17,407.40)(49.000,-80.403){2}{\rule{0.400pt}{0.385pt}}
\multiput(651.00,325.92)(0.911,-0.497){51}{\rule{0.826pt}{0.120pt}}
\multiput(651.00,326.17)(47.286,-27.000){2}{\rule{0.413pt}{0.400pt}}
\multiput(700.00,298.92)(2.133,-0.492){21}{\rule{1.767pt}{0.119pt}}
\multiput(700.00,299.17)(46.333,-12.000){2}{\rule{0.883pt}{0.400pt}}
\multiput(750.58,288.00)(0.498,1.733){95}{\rule{0.120pt}{1.480pt}}
\multiput(749.17,288.00)(49.000,165.929){2}{\rule{0.400pt}{0.740pt}}
\multiput(799.58,457.00)(0.498,2.051){95}{\rule{0.120pt}{1.733pt}}
\multiput(798.17,457.00)(49.000,196.404){2}{\rule{0.400pt}{0.866pt}}
\multiput(848.58,652.45)(0.498,-1.249){95}{\rule{0.120pt}{1.096pt}}
\multiput(847.17,654.73)(49.000,-119.725){2}{\rule{0.400pt}{0.548pt}}
\multiput(897.58,529.77)(0.498,-1.456){97}{\rule{0.120pt}{1.260pt}}
\multiput(896.17,532.38)(50.000,-142.385){2}{\rule{0.400pt}{0.630pt}}
\multiput(947.58,385.72)(0.498,-1.167){95}{\rule{0.120pt}{1.031pt}}
\multiput(946.17,387.86)(49.000,-111.861){2}{\rule{0.400pt}{0.515pt}}
\multiput(996.00,274.93)(3.696,-0.485){11}{\rule{2.900pt}{0.117pt}}
\multiput(996.00,275.17)(42.981,-7.000){2}{\rule{1.450pt}{0.400pt}}
\multiput(1045.58,269.00)(0.498,1.034){95}{\rule{0.120pt}{0.924pt}}
\multiput(1044.17,269.00)(49.000,99.081){2}{\rule{0.400pt}{0.462pt}}
\multiput(1094.00,368.92)(0.509,-0.498){95}{\rule{0.508pt}{0.120pt}}
\multiput(1094.00,369.17)(48.945,-49.000){2}{\rule{0.254pt}{0.400pt}}
\multiput(1144.00,319.92)(0.544,-0.498){87}{\rule{0.536pt}{0.120pt}}
\multiput(1144.00,320.17)(47.888,-45.000){2}{\rule{0.268pt}{0.400pt}}
\multiput(1193.58,276.00)(0.498,0.910){95}{\rule{0.120pt}{0.827pt}}
\multiput(1192.17,276.00)(49.000,87.284){2}{\rule{0.400pt}{0.413pt}}
\multiput(1242.58,365.00)(0.498,1.897){95}{\rule{0.120pt}{1.610pt}}
\multiput(1241.17,365.00)(49.000,181.658){2}{\rule{0.400pt}{0.805pt}}
\multiput(1291.58,550.00)(0.498,2.201){97}{\rule{0.120pt}{1.852pt}}
\multiput(1290.17,550.00)(50.000,215.156){2}{\rule{0.400pt}{0.926pt}}
\multiput(1341.00,769.59)(3.696,0.485){11}{\rule{2.900pt}{0.117pt}}
\multiput(1341.00,768.17)(42.981,7.000){2}{\rule{1.450pt}{0.400pt}}
\multiput(1390.58,772.64)(0.498,-0.890){95}{\rule{0.120pt}{0.810pt}}
\multiput(1389.17,774.32)(49.000,-85.318){2}{\rule{0.400pt}{0.405pt}}
\put(60,276){\circle{12}}
\put(109,372){\circle{12}}
\put(159,539){\circle{12}}
\put(208,659){\circle{12}}
\put(257,424){\circle{12}}
\put(306,327){\circle{12}}
\put(356,276){\circle{12}}
\put(405,294){\circle{12}}
\put(454,338){\circle{12}}
\put(503,451){\circle{12}}
\put(553,477){\circle{12}}
\put(602,409){\circle{12}}
\put(651,327){\circle{12}}
\put(700,300){\circle{12}}
\put(750,288){\circle{12}}
\put(799,457){\circle{12}}
\put(848,657){\circle{12}}
\put(897,535){\circle{12}}
\put(947,390){\circle{12}}
\put(996,276){\circle{12}}
\put(1045,269){\circle{12}}
\put(1094,370){\circle{12}}
\put(1144,321){\circle{12}}
\put(1193,276){\circle{12}}
\put(1242,365){\circle{12}}
\put(1291,550){\circle{12}}
\put(1341,769){\circle{12}}
\put(1390,776){\circle{12}}
\put(1439,689){\circle{12}}
\put(310,820){\circle{12}}
\put(80,779){\makebox(0,0)[l]{segments}}
\put(260.0,779.0){\rule[-0.200pt]{24.090pt}{0.400pt}}
\put(260.0,769.0){\rule[-0.200pt]{0.400pt}{4.818pt}}
\put(360.0,769.0){\rule[-0.200pt]{0.400pt}{4.818pt}}
\put(60.0,82.0){\rule[-0.200pt]{0.400pt}{175.616pt}}
\put(50.0,82.0){\rule[-0.200pt]{4.818pt}{0.400pt}}
\put(50.0,811.0){\rule[-0.200pt]{4.818pt}{0.400pt}}
\put(356.0,82.0){\rule[-0.200pt]{0.400pt}{70.343pt}}
\put(346.0,82.0){\rule[-0.200pt]{4.818pt}{0.400pt}}
\put(346.0,374.0){\rule[-0.200pt]{4.818pt}{0.400pt}}
\put(750.0,82.0){\rule[-0.200pt]{0.400pt}{67.211pt}}
\put(740.0,82.0){\rule[-0.200pt]{4.818pt}{0.400pt}}
\put(740.0,361.0){\rule[-0.200pt]{4.818pt}{0.400pt}}
\put(1045.0,82.0){\rule[-0.200pt]{0.400pt}{58.780pt}}
\put(1035.0,82.0){\rule[-0.200pt]{4.818pt}{0.400pt}}
\put(1035.0,326.0){\rule[-0.200pt]{4.818pt}{0.400pt}}
\put(1193.0,82.0){\rule[-0.200pt]{0.400pt}{71.306pt}}
\put(1183.0,82.0){\rule[-0.200pt]{4.818pt}{0.400pt}}
\put(60,82){\makebox(0,0){$+$}}
\put(356,82){\makebox(0,0){$+$}}
\put(750,82){\makebox(0,0){$+$}}
\put(1045,82){\makebox(0,0){$+$}}
\put(1193,82){\makebox(0,0){$+$}}
\put(310,779){\makebox(0,0){$+$}}
\put(1183.0,378.0){\rule[-0.200pt]{4.818pt}{0.400pt}}
\put(60.0,82.0){\rule[-0.200pt]{332.201pt}{0.400pt}}
\put(1439.0,82.0){\rule[-0.200pt]{0.400pt}{187.420pt}}
\put(60.0,860.0){\rule[-0.200pt]{332.201pt}{0.400pt}}
\put(60.0,82.0){\rule[-0.200pt]{0.400pt}{187.420pt}}
\put(750,0){\makebox(0,0){segment index}}%
\put(20,500){%
\special{ps: gsave currentpoint currentpoint translate
270 rotate neg exch neg exch translate}%
\makebox(0,0)[b]{\shortstack{similarity}}%
\special{ps: currentpoint grestore moveto}%
}%
\end{picture}
  \caption[A smoothed segmentation.]{A smoothed segmentation. (window size: 25, average real segment length: 240)}
  \label{fig:seg:segFinal}
\end{figure}
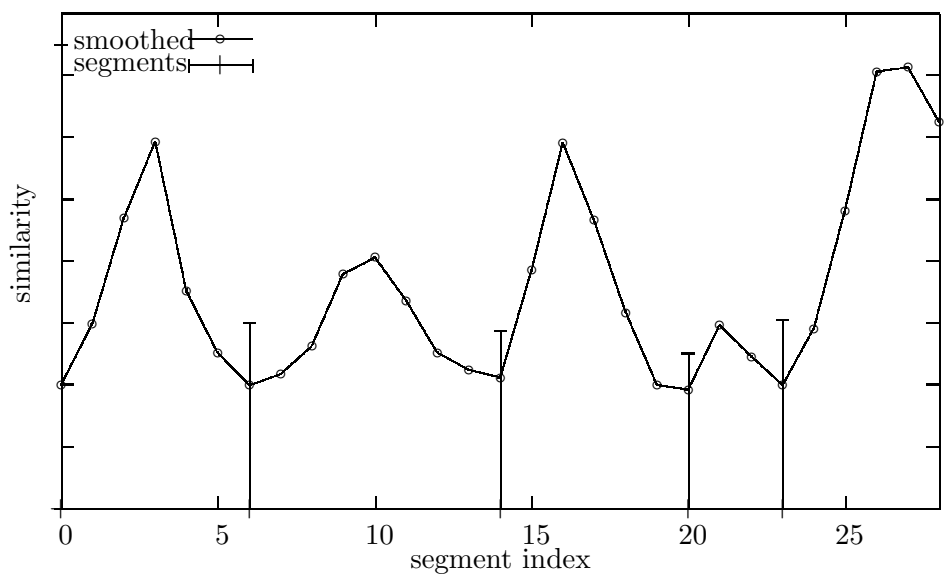

The boundaries for the text can be extracted from the obtained curve
by detecting the local minima. To avoid the local perturbations that
could have got through the smoothing process, each boundary is given a
relevance value. The relevance of the segmentation at the $i^{st}$
point of the curve, is the average of the last local maximum before
$i$ and the first local maximum after $i$. The figure
\ref{fig:seg:segFinal} is an example (with the same text as in the
figure \ref{fig:seg:nonsmooth}) of the final result. The boundary
detected on a minimum near to its surrounding maxima has a low
relevance.

This relevance computation can be used to filter out the extraneous
boundaries. This will render the algorithm more robust as a small
window size will be applicable to long texts -- which contains
segments where sub-topic shifts can occur but are not interesting.

\section{Sentence Detection}
\label{sec:seg:sentence-detection}

Often, the initial windowing, which determines the boundary's positions,
is not the best solution. Indeed, window end/start rarely corresponds to what a
human would choose as a boundary; punctuation in the text is
present as a hint to the reader on low level subject shift, but our
algorithm does not have the knowledge to decide if a point is there to
end a sentence or to separate a number's digits.

Sentence detection could be implemented using \gloss[word]{pm}; it
could then be used to determine the window size -- expressed
  in terms of sentences in place of words -- or at the end of the
processing to refine the boundary position.

\section{Sensibility to the Text Type}
\label{sec:seg:sens-text-type}

Two parameters has to be fixed to use this method:
\begin{itemize}
\item the window size,
\item the smoothing amount.
\end{itemize}
These two parameters are often difficult to fix automatically as they
depend on the text nature. However, they provide a good flexibility to the
technique.

The size of the text is highly determinant in the choice of the window
size. For long texts, it is better to fix a high window size
whereas, for small texts, a lower window size is more applicable.
There is more probability than in a small text, the topics will be
treated in a smaller number of words. Still, a really low window size
can give incorrect results even for small text, see section
\ref{sec:eval:segmentation} for exegesis of this issue.

This technique, is more relevant to expository texts where the
different topics are treated in exclusive blocks of text. Different
topics will rarely be mentioned in the same part of the text but will
be distributed amongst different sections. For example this report
does not mention \emph{segmentation} in the same section as \emph{topic
  extraction}. Moreover, in scientific or technical texts, it is more
probable that the author use the same -- technical -- term with
redundancy.

For narrative texts, the technique is less robust as the author often
avoid repetitions. In this kind of texts, it is also natural to mix
more than one topic in the same section of the text. Thus,
segmentation of this kind of text is less effective and avoiding the
detection of too much irrelevant boundaries implies the use of higher
window size and of important smoothing, which could drastically
decrease the algorithm \gloss[word]{recall}.

\chapter{Topic Extraction}
\label{cha:extraction}

Topic extraction can be regarded as an extended keyword extraction
task. Keyword extraction is based on \gloss[none]{statmeth}statistical
methods to identify the content bearing words; this technique is based
on Zipf's Law \cite{zipf32freq}.

Relatively frequent words are selected whereas too frequent words
(i.e. common words) and low frequency words (i.e. those not really
associated to the topic) are discarded as determined by Luhn's upper
and lower cut-off \cite{luhn58automatic}. This extraction process has
some limits because it only extracts words present in the document.
The \gloss[word]{bow} extension \cite{ferret98thematic} or the
\gloss[word]{lexicalchains} techniques \cite{barzilay97using} (see
section \ref{sec:state-art:topic-extraction}) propose some possible
solutions.  They use an external linguistic resource (i.e. an
\gloss[word]{ontology}) to take into account possible
\gloss[none]{hypernym}hypernyms of the document's keywords.

This project develops a novel method contrasting with the
\gloss[word]{bow} technique. The novel approach is to give importance
to the conceptual position in the \gloss[word]{ontology} during the
selection process.  When extending the \gloss[word]{bow}
representation of the document, the existing techniques introduce new
words and give them a weight in correspondence to their links with the
already present words. Our approach is slightly different, as the
initial \gloss[word]{bow} representation of the document is not
extended but the conceptual hierarchy induced by this representation
is constructed and concepts believed to be representative of the
document extracted from it.

The aim of this technique is to extract concepts in place of simple
words. Concepts provided by the \gloss[word]{ontology} are often
associated with a word and/or a gloss (see section
\ref{cha:edr-dict-ontol}) and are therefore more interesting. Indeed,
the extracted concepts will not necessarily be the leaf ones -- i.e.
the ones directly triggered by the words present in the document --
but more general concepts located higher in the hierarchy. Our
selection process tends to favor concepts which are occurs frequently
in the document -- because some of their sub-concepts are present in
the document -- and that present an acceptable genericity (see section
\ref{sec:cut-extraction}).

For example, the following text:
\begin{quotation}
  For his fourth birthday, Kevin was brought to the zoo with all his
  friends. The zoo was invaded by running children. The young child
  had never seen so many animals which in turn had never seen so many joyful
  children. During his visit, the child discovered strange cats that
  were called ``lion'' or ``tiger'' and that were quite frightening in
  spite of their fluffy appearance. The child also discovered the huge
  mammals: the giant giraffe, the Indian elephant and its African
  cousin.  Kevin and the other children were really impressed by the
  multicolor parrots and the giant ostrich.
  
  After the walk through the park, all the children were directed to
  the aquarium where they discovered small fish swimming with
  threatening sharks. Every child wanted to have a photo with their
  mechanical replica in the hall.
  
  At the end of the day, they all got an ice cream. Each child was a
  little crisis of its own when it was his turn to choose the flavor
  but at the end, all the children were contented and Kevin's
  birthday was another dream day in a child life.
\end{quotation}
will certainly raise the \textbf{child} lemma as a representative keyword,
because it is well present in the text in the form of ``child'' and
``children''. However, the important topics of \textbf{zoo} and
\textbf{animal} that are mentioned in the text only once or twice will
be discarded. However, if the words present were aggregated to more
generic concepts, all the mentioned animals will certainly raise the
concept \textbf{animal} whereas ``zoo'', ``park'' and ``aquarium'' could
raise a concept indicating an \textbf{amusement park}.

\section{Preprocessing}
\label{sec:ext:preprocessing}

Preprocessing of documents is an issue that has found more than one
solution. Two opposed directions are currently chosen in the text
mining field:
\begin{itemize}
\item either a shallow preprocessing is performed and the consecutive
  processing will perhaps perform with irrelevant data,
\item or the preprocessing is a complex task that drastically
  decreases the amount of data that can be processed.
\end{itemize}
The issue with preprocessing is that most of the data generated at the
creation of the document will only produce noise in the output of the
algorithm. However, there is no efficient way to know what will
produce noise and what is interesting in the original data. The first
approach avoids loosing too many interesting data in the
preprocessing; this is a good way of keeping the output
\gloss[word]{recall} high, but lowers the algorithm
\gloss[word]{precision}. The second takes the opposite way: the output
\gloss[word]{recall} decreases as some interesting data are lost but
the \gloss[word]{precision} is higher because the good results are not
drowned in the noise.

The first approach has been chosen. A trivial preprocessing (i.e.
\gloss[word]{lemmatization} and a \gloss[word]{stoplist}) is performed
to render the document representation compatible with the EDR database
and to remove the common words. Then, using an existing
\gloss[word]{pos} tagger trained on the EDR data (see section
\ref{sec:corp-transf}), \gloss[word]{pos} disambiguation is performed.

\subsection{Tokenization}
\label{sec:ext:token}

The initial step in many textual data mining algorithms is the
\gloss[word]{tokenization}. This technique is used to split the
document in distinct tokens that humans will recognize as words or
\gloss[word]{compounds}. In this project, the
\gloss[word]{tokenization} had to be performed with the intent of
retrieving the entry corresponding to the tokens from the EDR database.
More precisely, the available entry points in this dictionary are the
word's \gloss[none]{lemma}lemmas and this implies, in addition to the
\gloss[word]{tokenization}, the \gloss[word]{lemmatization} of the
document. Both steps rely on the use of a \gloss[word]{lexicon}
extracted from the EDR dictionary (see section
\ref{sec:lexicon-crea}).

\subsection*{Compound Word Lemmatization}

The main problem for standard tokenization is the processing of
\gloss[word]{compounds}.  Compound words are composed of more than one
word and it is difficult to know when it is correct and not to
aggregate multiple words to compose one. Complex techniques has been
developed to identify compounds in text\cite{shin97multivariate}.
However, our approach was simpler; because we need entry points in the
EDR, the compounds defined in the database were also included during
the lexicon creation process. Then, during the tokenization, the
tokens are identified as the longest character chain that can be
matched with an entry in the lexicon.  Hence, whenever it is possible
the longest compound is selected.

\subsection*{Making the Lemmatizer More Robust}

One obstacle encountered with this technique is that EDR provides some
\gloss[word]{compound} proper nouns; these words are either
capitalized on the first letter of the first word or on the first
letter of all words. While processing the documents, it's frequent to
encounter the same \gloss[word]{compounds} with a different
capitalization and it's not really efficient to try all possible
capitalizations. This issue is associated to the more general question
of the presence of proper nouns in the \gloss[word]{ontology}. Most of
the proper nouns encountered in a document are not present in EDR and
can therefore not be processed by our algorithm; and even if they are
present, they are attached to irrelevant concepts. For example, a
first name like Michael is attached to the concept: ``in the Bible,
the archangel who led a group of loyal angels against Satan''; which
is rarely what is meant in the document. In some cases, these nouns
are the real topics of the text and it is incorect not to take them
into account because they are absent from the resource used. Thus
implementation of one of the existing proper name identification
technique \cite{wacholder97disambiguation} would be optimal.

\begin{figure}[h]
  \hrule \smallskip I saw the man singing with girls in Central Park.
  The system output
  is ready.\\
  \hrule \smallskip I(Pron.)| saw(Verb)| the(Det.)| man(Noun)|
  sing(Verb)| with(?)| girl(Noun)| in(Adv.)| central(Adj.)|
  park(Noun)| the(Det.)| system output(Noun)| is(Verb)| ready(Adj.)
  \smallskip \hrule
  \caption{Example of a tokenization/lemmatization}
  \label{fig:ext:extok}
\end{figure}

Sadly, there was no simple solution to the capitalization and proper
noun identification. Thus, in this project, some interesting words are
discarded because the tokenizer is unable to identify a corresponding
entry in the database. This is often a cause of loss in
\gloss[word]{precision} -- when a proper name is not identified
correctly -- and \gloss[word]{recall} -- when a proper name is
completely discarded -- of the output. In other cases, simple lemmas
are identified in place of the real proper names. In the example
illustrated in figure \ref{fig:ext:extok}, ``system output'' which is
not capitalized is identified as a compound, whereas ``Central Park''
-- which is a proper name not present in EDR -- is only identified as
the individual lemmas ``central'' and ``park'', which do not retain
the proper name meaning.

\subsection{Part of Speech and Word Sense Disambiguation}
\label{sec:ext:disambiguation}

Disambiguation is one of the main goals of preprocessing (see
\cite{brill95pos} and \cite{veronis98wsd}): removing some
undesired meanings of the lemmas we have found. Indeed, a word alone
raises a lot of ambiguity in the algorithm:
\begin{itemize}
\item an inflection can be produced from more than one \gloss[word]{lemma}; ``left''
  can be the adjective ``left'' -- as opposed to right -- or the
  simple past of the verb ``leave''.
\item a \gloss[word]{lemma} can correspond to more than one \gloss[short]{pos}
  category; ``run'' could be a Noun or a Verb.
\item a \gloss[word]{lemma} with a defined \gloss[word]{pos} can represent more than
  one concept; a ``controller'' could be:
  \begin{itemize}
  \item a device which controls certain subjects,
  \item an executive or official who supervises financial affairs,
  \end{itemize}
\end{itemize}

There are no perfect solutions to these problems as explained before;
too much disambiguation would inevitably reduce the algorithm recall
and therefore its accuracy. The first problem is quite unsolvable
without any information on the word's context, however, a
\gloss[short]{pos} tagger can help in resolving this type of
ambiguities. The last ambiguity could be dealt with by using existing
technique, however, in this project, no word sense disambiguation was
implemented as it was out of it's scope.

\subsubsection{Part Of Speech Disambiguation}
\gloss[short]{pos} disambiguation corrects the second type of
ambiguities -- and in some cases the first ones.  \gloss[short]{pos}
tagging \cite{brill95pos} is a state of the art technique that is
known to provide very good performances. Using a machine learning
approach trained on a large \gloss[word]{corpus}, the tagger
can deduce the \gloss[short]{pos} category of each word in the
document.  In some cases, there is still some ambiguity, but the
choices are drastically reduced.

However, the tagger requires a large training \gloss[word]{corpus}
where each token is annotated with the correct \gloss[short]{pos} tag.
In this project, the \gloss[word]{corpus} used needs to be the same as
the one used to construct the EDR database as the tokenization must
correspond. An existing tagger \cite{treetagger} was trained on a
corpus adapted from the EDR data (see section \ref{sec:corp-transf})
but no evaluation was performed on the robustness of this training.

As explained in section \ref{sec:corp-transf}, a new tag set has been
chosen. Transformation from this new tag set to the EDR dictionary set
to get an entry point in the hierarchy should have been trivial, but
in EDR the dictionary tag set is different from the corpus one.
Hence, a transformation table was constructed. This
transformation table takes a lemma and its \gloss[short]{pos} -- which
are returned by the tagger -- to transform them in an entry in the
dictionary.

This is done by mapping the tagger \gloss[short]{pos} to all its
possible \emph{inflection pattern}/\emph{word form}/\emph{edr pos}
triplet in the EDR database. Then, a unique entry that has these
grammatical information and the lemma returned by the tagger can be
found in the database.

\subsubsection{Word Sense Disambiguation}
The third case of ambiguity is frequent in our processing, as
several concepts are associated to one \gloss[word]{lemma} in the EDR
database. This ambiguity is really hard to resolve during
the preprocessing, since there is not enough contextual information to
discern which sense is the best. The extraction algorithm is in fact
the one that should perform the conceptual disambiguation.

This process, known as Word Sense Disambiguation \cite{veronis98wsd},
is another leading research in the field of Natural Language
Processing. Identifying the correct concept attached to a word in the
document is a complex task that requires large linguistic resources
and time consuming algorithms. For example, one method
\cite{diab02unsupervised} -- proposed by M.Diab and P.Resnik -- uses
unsupervized learning on aligned bilingual corpora to perform
disambiguation. Word sense disambiguation is therefore a hard task
that requires fully dedicated algorithm and resources. Hence, WSD
disambiguation has been left apart for a future project.

\section{Document Representation}
\label{sec:chap:docum-repr}

The preprocessing step leaves the data in a state that is not
interesting for the subsequent treatments. The preliminary document
representation is a list of words and all the possible lemmas (with
\gloss[word]{pos} information) selected by the tagger. Each word in
the post-preprocessing representation holds information on its
position in the text, but no information on concepts they are linked
to in the EDR hierarchy.

As it is usual in the textual mining field, the first representation
computed from the document is a \gloss[word]{bow}. Then, using the
lemmas attached to each words, a \emph{bag of concepts} is constructed.
Each concept is attached to the lemmas that raised it, this one is
then attached to each word that raised the lemma. This could help in
the future if the relative frequency of each concept should be used in
the extraction algorithm.

\section{Topic Selection}

Once the initial conceptual representation of the document have been
constructed, the novel algorithm will construct the spanning
\gloss[word]{DAG} over the \emph{bag of concepts}. Each concept in
this hierarchy represents a part of the document, in a \emph{more or
  less generic} way.

The goal of the algorithm is to extract a set of concepts that will
represent the entire document. This set must retain the information
contained in the initial set of concepts but it should also provide a
more generic representation.  For example, all the leaves of the
hierarchy are a possible representation, but there is no genericity
gained with respect to the initial representation.  Another selection
in the hierarchy could be its root, this concept is the most generic
representation of the document, however, it does not retain anything
from the information contained in the document.

These two examples are in fact trivial cuts in a tree: the leaves or
the root. A cut in a tree, is a minimal set of nodes that covers all
the leaves. The set being minimal, it guarantees that it does not
contain a node and one of its subordinates -- i.e the cut is not made
of nodes that are subordinate with each other.

A cut in our hierarchy would then be a good selection of concepts to
represent the content of the document. However, the extracted
hierarchy is a \gloss[short]{DAG} and cannot guarantee -- because of
multiple inheritance -- that a concept in a cut will not be a
subordinate of another concept in this cut (see algorithm
\ref{alg:sec:cut-extraction}). As discussed in section
\ref{sec:new-tables}, if a concept has multiple inheritance, it is
not considered as having the same meaning.

Then it is acceptable to have a concept C that subordinates a concept
A and B and choose A and C in the cut. If the concept C is selected,
it will happen because of its inheritance from B that have a different
meaning than if it was selected as the subordinate of A.

\subsection{Cut Extraction}
\label{sec:cut-extraction}

The approach of this algorithm is to extract a cut in the hierarchy
that will represent our document. Cuts in the hierarchy are scored
and the best one is selected. However, in a regular tree with a
branching factor $b$, the number of cut of depth $p$ is defined by:
\begin{align*}
  \mathrm{C}(p=1) = 1\\
  \mathrm{C}(p>1) = \mathrm{C}(p-1)^b + 1
\end{align*}
The number of cut is then exponential in the depth of the tree and
evaluating the scores of all cuts in the extracted hierarchy is
impossible for a real time algorithm. A dynamic programming algorithm,
presented in \ref{alg:sec:cut-extraction} was developed to avoid
intractable computations.

\begin{algorithm}
\centering
  \caption{Cut Extraction Algorithm}
  \label{alg:sec:cut-extraction}
  \begin{algorithmic}[3]
  \STATE initiate the \textbf{actual} array with the leaves
  \WHILE{\textbf{actual} is not empty}
  \FORALL{concept \textbf{C} in \textbf{actual}}
  \STATE add every superconcepts of \textbf{C} to array \textbf{next}
  \STATE \textbf{L} $\leftarrow$ concept local score
  \STATE \textbf{G} $\leftarrow$ average of \textbf{C} subconcepts score
  \IF{$\textbf{L} \leq \textbf{G}$}
  \STATE store \textbf{G} as the score of \textbf{C}
  \STATE mark that \textbf{C} should be expended
  \ELSE
  \STATE store \textbf{L} as the score of \textbf{C}
  \STATE mark that \textbf{C} should not be expended
  \ENDIF
  \ENDFOR
  \STATE copy \textbf{next} to \textbf{actual}
  \ENDWHILE

  \STATE push the root concept on the \textbf{stack}
  \WHILE{\textbf{stack} is not empty}
  \STATE \textbf{C} $\leftarrow$ pop(\textbf{stack)}
  \IF{\textbf{C} should be extended}
  \STATE push all leaves on \textbf{stack}
  \ELSE
  \STATE select \textbf{C}
  \ENDIF
  \ENDWHILE
  \end{algorithmic}
\end{algorithm}

One cut is a representation of the leaves of the tree. In this
context, the score of a cut $\chi$ is computed relatively to the
leaves covered. Let $\{f_1, f_2, \ldots, f_M\}$ be the set of leaves
in the extracted hierarchy -- i.e. the initial representation covered
by every cut -- and $c_i$ the concept that covers the leaf $f_i$ in
the actual cut. A score $\mathrm{U}(c_i)$ is computed for each
$(c_i,f_i)$ pair to measure how much the concept $c_i$ is
representative of the leaf $f_i$ in the cut $\chi$ (see next section).

An intuitive score, selected for our algorithm, is the average over
all leaves of the score given to each concept in the cut:
$$
\mathrm{S}(\chi) = \frac{1}{M}\times \sum_{i=1}^{M} \mathrm{U}(c_i,\chi)
$$

\subsection{Concept Scoring}
\label{sec:concept-scoring}

For each concept, a score measuring how much it represents the leaves
it covers in the initial \emph{bag of concepts} must be computed. As
explained before, this score should reflect the genericity of a
concept compared to the leaves it covers, but should also take into
account the amount of information that is kept about these leaves in
the selected concept.

\subsubsection{Genericity}
\label{sec:genericity}

It is quite intuitive that, in a conceptual hierarchy, a concept is
more generic than its subconcepts. The higher we are in the hierarchy,
the less specific the concept will be. For example, the concept for
``animal'' is less specific than the concept representing ``dog''.

In the same time, even if we are in a \gloss[short]{DAG}, the higher a
concept is in the hierarchy, the higher is the number of covered
leaves. For example, if ``cat'', ``dog'' and ``snake'' are leaves of
the hierarchy, the concept ``vertebrate'' will cover ``cat'' and
``dog'' whereas, it's superconcept ``animal'' will cover ``cat'',
``dog'' and ``snake''. Following this assertion, a simple score $S_1$ have
been chosen to describe the genericity level of a concept. This score
will be proportional to the total number of covered leaves $n_i$ for a
concept $i$:

$$S_1(i,\chi) = \mathrm{f}(n_i)$$

Two constraints are introduced:
\begin{itemize}
\item $\mathrm{f}(1) = 0$, a concept covering one leaf is not more
  generic than that leaf,
\item $\mathrm{f}(N) = 1$, the root, covering all leaves, is the most
  generic concept ($N$ being the total number of leaves in the
  hierarchy).
\end{itemize}

With a linear function of the form $\mathrm{f}(x) = a.x + b$, the
score can be written:

$$S_1(i,\chi) = \mathrm{f}(n_i) = \frac{n_i-1}{N-1}$$

\subsubsection{Informativeness}
\label{sec:doc-dist}

Even if the algorithm should select generic concepts, it cannot always
choose the most generic ones, as it will always be the root of the
hierarchy. The algorithm, has to take into account the amount of
information kept by the selected concepts about the document.

Each concept represents, in a cut, a certain number of leaves of the
initial representation. If the concept that is selected to represent a
leaf is the leaf itself, then no information is lost. In the same way,
if the root is selected to describe a leaf, then all the semantic
information retained by this leaf will be lost -- i.e. the root
represents any leaf in the same way.

An intuitive measure to describe such behavior would be to compute the
distance -- in number of edges -- between the concept $i$ and the leaf
it is supposed to represent in the cut $\chi$. A second score $S_2$ is
defined for each concept in the cut. The two constraints exposed above
are introduced:
\begin{itemize}
\item $S_2(leaf,\chi) = 1$ a leaf represents perfectly itself,
\item $S_2(root,\chi) = 0$ the root does not describe anything.
\end{itemize}

As discussed in section \ref{sec:cptscore}, a concept covers more than
one leaf. Let $N$ be the number of path to the root, $n$ the number of
covered leaves (see figure \ref{fig:edr:path}) in the document
hierarchy. The average normalized separation between a concept and its
leaves can then be defined as:

$$
d(i) = \frac{1}{N \times n} \times \sum_{j=0}^{N}\sum_{k=0}^{n} (\frac{l_{ij}}{L_{ij}+l_{ik}})
$$

This measure is equal to 1 for the root and 0 for all leaves.
Therefore, $S_2$ can be computed with:

$$S_2(i,\chi) = 1 - d(i)$$

\subsubsection{Score Combination}
\label{sec:score-combination}

Two scores are computed for each concept in the evaluated cut,
however, a unique score $\mathrm{U}(c_i)$ is needed in the cut scoring
scheme. Therefore, a combination formula was chosen for $S_1$ and
$S_2$. A weighted geometric mean provides a way to render the score
independent to the summing formula used in the cut scoring scheme:

$$
\mathrm{U}(c_i,\chi) = S_1(i,\chi)^{1-a} \times S_2(i,\chi)^{a}
$$

The parameter $a$ is meant to offer a control over the number of
concepts returned by the extraction algorithm. If the value of $a$ is
near to one, then it will advantage the score $S_2$ over $S_1$,
and the algorithm will prefer to extract a cut near to the
leaves, which will inevitably contain a greater number of concepts.
Whereas a value near zero will advantage $S_1$ over $S_2$ and
therefore prefer more generic concepts in the cut which will be more
compact.

\chapter{Evaluation}
\label{cha:evaluation}

Evaluating the whole system, i.e. performing segmentation and then
extraction, would require a testing corpus with complex annotations. A
set of texts with segments and their corresponding topics should be
used; however, no such corpus is freely available. The evaluation of
this project has hence been separated in two different steps;
evaluation of the segmentation and evaluation of the topic extraction
separately.

\section{Evaluation Corpora}
\label{sec:evaluation-corpora}

\subsection{Reuters Corpus}
\label{sec:eval:reuters-corpus}

Segmentation evaluations have been made on the Reuters
corpus\footnote{http://about.reuters.com/researchandstandards/corpus/}.
The Reuters corpus is made of 806'791 pieces of news, including all
English Reuters' news edited between 20/08/1996 and 19/08/1997. The
important characteristic of this corpus is that it is freely available
to the research community while having a professional content and
meta-content.

All news are stored in a separate \gloss[short]{XML} file presenting
rich information about the content of the news. Annotation is given
about included topics, mentioned countries and industries, for all of
the news. These annotations were either automatically or hand checked
and corrected to produce an accurate content. Topic codes and industry
codes are organized in a hierarchy to produce more complete
information. 

The annotation is a limited set of topics that features more
information on the text category than on its discussed topics.
Therefore, it was not used in evaluating the extraction algorithm but,
as it provides a quantity of small stories that can be concatenated,
it has been used to test the segmentation.

Here is an example of an entry:

\begin{quote}
  Europe's leading car-maker Volkswagen AG insists the hotly debated
  subsidies it received to invest in former communist east Germany
  were legal, the Financial Times said in its weekend edition.
  
  The newspaper quoted Volkswagen chairman Ferdinand Piech as saying
  the group was convinced the funds were legal and would stick to its
  investment plans in the state of Saxony.
  
  Piech said: We received it (the investment) legally and because we
  did so, we will invest it correctly. We had the choice of any
  location: without support, it would not have been this one.
  
  A row over the funds erupted last month when Kurt Biedenkopf, state
  premier of the east German state of Saxony, paid out 142 million
  marks to Volkswagen, of which Brussels says 91 million marks were
  unauthorized.
  
  The Saxony payment followed a European ruling in late June clearing
  only 540 million marks in east German subsidies for the carmaker out
  of an initial plan for 780 million.
  
  Volkswagen had threatened to pull the plug on its plans to build
  plants in the Saxony towns of Mosel and Cheimnitz if the funds were
  not paid in full.
  
  Saxony, which has a jobless rate of more than 15 percent, is
  determined to retain the car giant's interest in the region.
  
  European Competition Commissioner Karel Van Miert and German
  Economics Minister Guenter Rexrodt met in Brussels on Friday to
  resolve the issue but could agree only to continue their efforts to
  find a solution.
  
  The Commission is due to discuss the case on September 4.
  
  Bonn and the Commission are to continue their talks.
  
  TOPICS: STRATEGY/PLANS and FUNDING/CAPITAL.
\end{quote}

\subsection{INSPEC Bibliographic Database}
\label{sec:inspec-corpus}

\subsubsection{Description}

INSPEC\footnote{http://www.iee.org/publish/inspec/} is a database of
bibliographic resources about physics, electronics and computing and
is composed of:
\begin{itemize}
\item physics abstracts,
\item electrical \& electronics abstracts,
\item computer \& control abstracts,
\item business automation.
\end{itemize}

A set of 238 abstracts were randomly extracted from this database to
create a base for the topics extraction evaluation process. In the
database, each abstract is annotated with two set that are interesting
for the evaluation:
\begin{itemize}
\item[ID] Key Phrase Identifiers, containing words assigned by an
  indexer. These give a description in an open language form of the
  content of the document.
\item[DE] Descriptors, which is a set of keywords describing the
  content of the document.
\end{itemize}

The corpus created is then composed of small texts -- i.e. the
abstracts -- annotated by a set of free language words that can be
found in the document, but not imperatively in the abstract.
Therefore, it provides a good evaluation base for the extraction
algorithm as this one is meant to extract topics that are not required
to be found in the text. Here is an example of an entry:

\begin{itemize}
\item[AB] The entire collection of about 11.5 million MEDLINE
  abstracts was processed to extract 549 million noun phrases using a
  shallow syntactic parser. English language strings in the 2002 and
  2001 releases of the UMLS Metathesaurus were then matched against
  these phrases using flexible matching techniques. 34\% of the
  Metathesaurus names occurring in 30\% of the concepts were found in
  the titles and abstracts of articles in the literature. The matching
  concepts are fairly evenly chemical and non-chemical in nature and
  span a wide spectrum of semantic types. This paper details the
  approach taken and the results of the
  analysis.
\item[DE] knowledge-representation-languages;
  medical-information-systems; text-analysis;
\item[ID] UMLS-Metathesaurus; MEDLINE-; shallow-syntactic-parser;
  flexible-matching-techniques
\end{itemize}

\subsubsection{Transformation}

The extraction algorithm gives as its output a list of concepts in the
EDR dictionary. Therefore, the list of words describing the document's
content -- given in an open language form -- cannot be used directly
for the evaluation. For each word associated to the abstract, a list
of concept have been extracted from EDR.

In this transformation process, either there exists at least one
concept in EDR that is directly raised by the word or there is no
concept in EDR that describes this word. There can be more than one
concept associated to a word -- due to multiple word senses -- but
there is no way to know which one is meant in the annotation,
therefore, all the possible senses are chosen. If there is no concept
associated to a word and its a compound word, then this one is cut in
smaller words using the tokenizer developed for EDR (see section
\ref{sec:ext:token}) and the concepts attached to all the compound
components are kept.

The corpus created in this process provides a set of small text entry
annotated with two sets of concepts:
\begin{itemize}
\item the ones directly triggered by a word in the corpus
  annotation,
\item the ones that can be deduced from the compound word
  segmentation.
\end{itemize}
The second set of concept has been left aside for the current
evaluation, as it is possible that it will add too much noise in the
evaluation. For example, if the word ``flexible matching techniques''
is segmented in the concept representing ``flexible'', ``matching''
and ``techniques'', there will be a lot of noise added by the multiple
senses of each word separately -- the meaning of the compound is not
the sum of all the meanings of its composing words.

\begin{figure}[htbp]
  \centering \setlength{\unitlength}{0.240900pt}
\ifx\plotpoint\undefined\newsavebox{\plotpoint}\fi
\sbox{\plotpoint}{\rule[-0.200pt]{0.400pt}{0.400pt}}%
\begin{picture}(1500,900)(0,0)
\sbox{\plotpoint}{\rule[-0.200pt]{0.400pt}{0.400pt}}%
\put(201.0,123.0){\rule[-0.200pt]{4.818pt}{0.400pt}}
\put(181,123){\makebox(0,0)[r]{ 0.2}}
\put(1419.0,123.0){\rule[-0.200pt]{4.818pt}{0.400pt}}
\put(201.0,197.0){\rule[-0.200pt]{4.818pt}{0.400pt}}
\put(181,197){\makebox(0,0)[r]{ 0.25}}
\put(1419.0,197.0){\rule[-0.200pt]{4.818pt}{0.400pt}}
\put(201.0,270.0){\rule[-0.200pt]{4.818pt}{0.400pt}}
\put(181,270){\makebox(0,0)[r]{ 0.3}}
\put(1419.0,270.0){\rule[-0.200pt]{4.818pt}{0.400pt}}
\put(201.0,344.0){\rule[-0.200pt]{4.818pt}{0.400pt}}
\put(181,344){\makebox(0,0)[r]{ 0.35}}
\put(1419.0,344.0){\rule[-0.200pt]{4.818pt}{0.400pt}}
\put(201.0,418.0){\rule[-0.200pt]{4.818pt}{0.400pt}}
\put(181,418){\makebox(0,0)[r]{ 0.4}}
\put(1419.0,418.0){\rule[-0.200pt]{4.818pt}{0.400pt}}
\put(201.0,491.0){\rule[-0.200pt]{4.818pt}{0.400pt}}
\put(181,491){\makebox(0,0)[r]{ 0.45}}
\put(1419.0,491.0){\rule[-0.200pt]{4.818pt}{0.400pt}}
\put(201.0,565.0){\rule[-0.200pt]{4.818pt}{0.400pt}}
\put(181,565){\makebox(0,0)[r]{ 0.5}}
\put(1419.0,565.0){\rule[-0.200pt]{4.818pt}{0.400pt}}
\put(201.0,639.0){\rule[-0.200pt]{4.818pt}{0.400pt}}
\put(181,639){\makebox(0,0)[r]{ 0.55}}
\put(1419.0,639.0){\rule[-0.200pt]{4.818pt}{0.400pt}}
\put(201.0,713.0){\rule[-0.200pt]{4.818pt}{0.400pt}}
\put(181,713){\makebox(0,0)[r]{ 0.6}}
\put(1419.0,713.0){\rule[-0.200pt]{4.818pt}{0.400pt}}
\put(201.0,786.0){\rule[-0.200pt]{4.818pt}{0.400pt}}
\put(181,786){\makebox(0,0)[r]{ 0.65}}
\put(1419.0,786.0){\rule[-0.200pt]{4.818pt}{0.400pt}}
\put(201.0,860.0){\rule[-0.200pt]{4.818pt}{0.400pt}}
\put(181,860){\makebox(0,0)[r]{ 0.7}}
\put(1419.0,860.0){\rule[-0.200pt]{4.818pt}{0.400pt}}
\put(201.0,123.0){\rule[-0.200pt]{0.400pt}{4.818pt}}
\put(201,82){\makebox(0,0){ 0}}
\put(201.0,840.0){\rule[-0.200pt]{0.400pt}{4.818pt}}
\put(356.0,123.0){\rule[-0.200pt]{0.400pt}{4.818pt}}
\put(356,82){\makebox(0,0){ 0.5}}
\put(356.0,840.0){\rule[-0.200pt]{0.400pt}{4.818pt}}
\put(511.0,123.0){\rule[-0.200pt]{0.400pt}{4.818pt}}
\put(511,82){\makebox(0,0){ 1}}
\put(511.0,840.0){\rule[-0.200pt]{0.400pt}{4.818pt}}
\put(665.0,123.0){\rule[-0.200pt]{0.400pt}{4.818pt}}
\put(665,82){\makebox(0,0){ 1.5}}
\put(665.0,840.0){\rule[-0.200pt]{0.400pt}{4.818pt}}
\put(820.0,123.0){\rule[-0.200pt]{0.400pt}{4.818pt}}
\put(820,82){\makebox(0,0){ 2}}
\put(820.0,840.0){\rule[-0.200pt]{0.400pt}{4.818pt}}
\put(975.0,123.0){\rule[-0.200pt]{0.400pt}{4.818pt}}
\put(975,82){\makebox(0,0){ 2.5}}
\put(975.0,840.0){\rule[-0.200pt]{0.400pt}{4.818pt}}
\put(1130.0,123.0){\rule[-0.200pt]{0.400pt}{4.818pt}}
\put(1130,82){\makebox(0,0){ 3}}
\put(1130.0,840.0){\rule[-0.200pt]{0.400pt}{4.818pt}}
\put(1284.0,123.0){\rule[-0.200pt]{0.400pt}{4.818pt}}
\put(1284,82){\makebox(0,0){ 3.5}}
\put(1284.0,840.0){\rule[-0.200pt]{0.400pt}{4.818pt}}
\put(1439.0,123.0){\rule[-0.200pt]{0.400pt}{4.818pt}}
\put(1439,82){\makebox(0,0){ 4}}
\put(1439.0,840.0){\rule[-0.200pt]{0.400pt}{4.818pt}}
\put(201.0,123.0){\rule[-0.200pt]{298.234pt}{0.400pt}}
\put(1439.0,123.0){\rule[-0.200pt]{0.400pt}{177.543pt}}
\put(201.0,860.0){\rule[-0.200pt]{298.234pt}{0.400pt}}
\put(201.0,123.0){\rule[-0.200pt]{0.400pt}{177.543pt}}
\put(40,491){\makebox(0,0){\begin{sideways}F-Measure\end{sideways}}}
\put(820,21){\makebox(0,0){concepts/keywords}}
\put(201,150){\usebox{\plotpoint}}
\multiput(201.58,150.00)(0.497,2.489){59}{\rule{0.120pt}{2.074pt}}
\multiput(200.17,150.00)(31.000,148.695){2}{\rule{0.400pt}{1.037pt}}
\multiput(232.58,303.00)(0.497,2.441){59}{\rule{0.120pt}{2.035pt}}
\multiput(231.17,303.00)(31.000,145.775){2}{\rule{0.400pt}{1.018pt}}
\multiput(263.58,453.00)(0.497,1.200){59}{\rule{0.120pt}{1.055pt}}
\multiput(262.17,453.00)(31.000,71.811){2}{\rule{0.400pt}{0.527pt}}
\multiput(294.58,527.00)(0.497,1.005){59}{\rule{0.120pt}{0.900pt}}
\multiput(293.17,527.00)(31.000,60.132){2}{\rule{0.400pt}{0.450pt}}
\multiput(325.58,589.00)(0.497,0.662){59}{\rule{0.120pt}{0.629pt}}
\multiput(324.17,589.00)(31.000,39.694){2}{\rule{0.400pt}{0.315pt}}
\multiput(356.58,630.00)(0.497,0.597){59}{\rule{0.120pt}{0.577pt}}
\multiput(355.17,630.00)(31.000,35.802){2}{\rule{0.400pt}{0.289pt}}
\multiput(387.00,667.58)(0.574,0.497){51}{\rule{0.559pt}{0.120pt}}
\multiput(387.00,666.17)(29.839,27.000){2}{\rule{0.280pt}{0.400pt}}
\multiput(418.00,694.58)(0.534,0.497){55}{\rule{0.528pt}{0.120pt}}
\multiput(418.00,693.17)(29.905,29.000){2}{\rule{0.264pt}{0.400pt}}
\multiput(449.00,723.59)(1.776,0.489){15}{\rule{1.478pt}{0.118pt}}
\multiput(449.00,722.17)(27.933,9.000){2}{\rule{0.739pt}{0.400pt}}
\multiput(480.58,732.00)(0.497,0.531){59}{\rule{0.120pt}{0.526pt}}
\multiput(479.17,732.00)(31.000,31.909){2}{\rule{0.400pt}{0.263pt}}
\multiput(511.00,765.58)(0.945,0.494){29}{\rule{0.850pt}{0.119pt}}
\multiput(511.00,764.17)(28.236,16.000){2}{\rule{0.425pt}{0.400pt}}
\multiput(541.00,781.60)(4.429,0.468){5}{\rule{3.200pt}{0.113pt}}
\multiput(541.00,780.17)(24.358,4.000){2}{\rule{1.600pt}{0.400pt}}
\multiput(572.00,785.59)(2.013,0.488){13}{\rule{1.650pt}{0.117pt}}
\multiput(572.00,784.17)(27.575,8.000){2}{\rule{0.825pt}{0.400pt}}
\put(603,793.17){\rule{6.300pt}{0.400pt}}
\multiput(603.00,792.17)(17.924,2.000){2}{\rule{3.150pt}{0.400pt}}
\multiput(634.00,795.60)(4.429,0.468){5}{\rule{3.200pt}{0.113pt}}
\multiput(634.00,794.17)(24.358,4.000){2}{\rule{1.600pt}{0.400pt}}
\multiput(665.00,799.61)(13.635,0.447){3}{\rule{8.367pt}{0.108pt}}
\multiput(665.00,798.17)(44.635,3.000){2}{\rule{4.183pt}{0.400pt}}
\put(727,802.17){\rule{6.300pt}{0.400pt}}
\multiput(727.00,801.17)(17.924,2.000){2}{\rule{3.150pt}{0.400pt}}
\put(758,803.67){\rule{7.468pt}{0.400pt}}
\multiput(758.00,803.17)(15.500,1.000){2}{\rule{3.734pt}{0.400pt}}
\multiput(789.00,805.61)(6.714,0.447){3}{\rule{4.233pt}{0.108pt}}
\multiput(789.00,804.17)(22.214,3.000){2}{\rule{2.117pt}{0.400pt}}
\multiput(820.00,808.60)(4.429,0.468){5}{\rule{3.200pt}{0.113pt}}
\multiput(820.00,807.17)(24.358,4.000){2}{\rule{1.600pt}{0.400pt}}
\multiput(851.00,812.61)(6.714,0.447){3}{\rule{4.233pt}{0.108pt}}
\multiput(851.00,811.17)(22.214,3.000){2}{\rule{2.117pt}{0.400pt}}
\put(882,814.67){\rule{7.468pt}{0.400pt}}
\multiput(882.00,814.17)(15.500,1.000){2}{\rule{3.734pt}{0.400pt}}
\put(913,815.67){\rule{14.936pt}{0.400pt}}
\multiput(913.00,815.17)(31.000,1.000){2}{\rule{7.468pt}{0.400pt}}
\put(975,817.17){\rule{18.700pt}{0.400pt}}
\multiput(975.00,816.17)(54.187,2.000){2}{\rule{9.350pt}{0.400pt}}
\put(1068,819.17){\rule{18.500pt}{0.400pt}}
\multiput(1068.00,818.17)(53.602,2.000){2}{\rule{9.250pt}{0.400pt}}
\put(1160,820.67){\rule{14.936pt}{0.400pt}}
\multiput(1160.00,820.17)(31.000,1.000){2}{\rule{7.468pt}{0.400pt}}
\put(1222,821.67){\rule{7.468pt}{0.400pt}}
\multiput(1222.00,821.17)(15.500,1.000){2}{\rule{3.734pt}{0.400pt}}
\put(1253,823.17){\rule{12.500pt}{0.400pt}}
\multiput(1253.00,822.17)(36.056,2.000){2}{\rule{6.250pt}{0.400pt}}
\put(201.0,123.0){\rule[-0.200pt]{298.234pt}{0.400pt}}
\put(1439.0,123.0){\rule[-0.200pt]{0.400pt}{177.543pt}}
\put(201.0,860.0){\rule[-0.200pt]{298.234pt}{0.400pt}}
\put(201.0,123.0){\rule[-0.200pt]{0.400pt}{177.543pt}}
\end{picture}
  \caption{F-measure against keywords on found concepts ratio}
  \label{fig:f-ratio}
\end{figure}
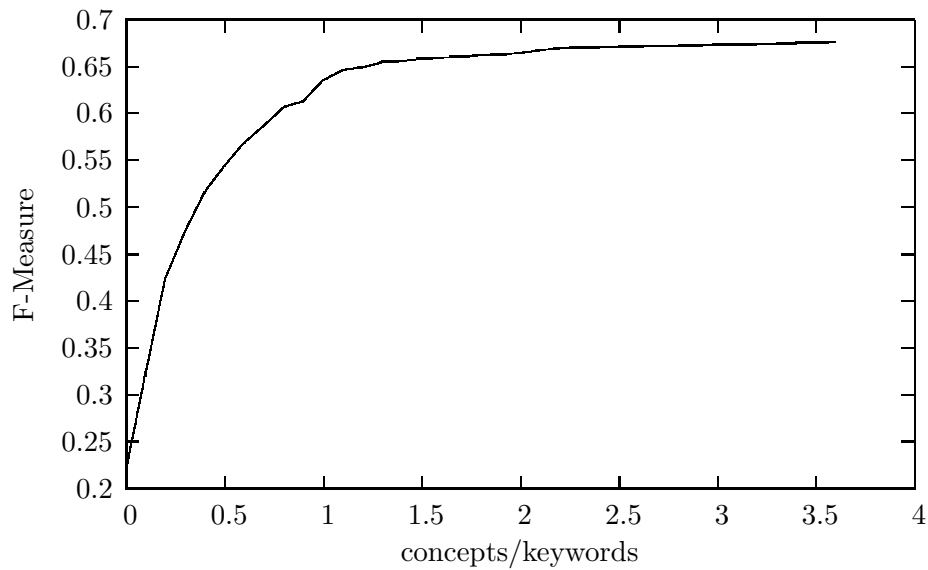

For some entries, the ratio between the number of words in the initial
annotation and the number of found concepts is two small. Therefore,
there is no guarantee that the low precision of the extraction
algorithm is generated by a bad extraction or the lack of translated
annotation (see figure \ref{fig:f-ratio}). Therefore, the corpus has been limited to
the entries that had a good ratio -- i.e. in out case, over 0.5.

\section{Segmentation}
\label{sec:eval:segmentation}

Segmentation was evaluated with a simple metric on approximately 2'910
texts. These texts were created by concatenating random sized news
from the Reuters corpus. A set of either 5 -- for 1'608 texts -- or 10
-- for 1'302 texts -- news were selected and composed the \emph{real
  segments} of the texts.

Evaluating the boundaries found by the automatic segmentation is then a
matter of comparing them with the news segments that were used. This
is not an obvious task as it is difficult to know the alignment
between the \emph{real segments} and the \emph{found} ones. The problem is
even more difficult to solve if the number of \emph{found segments} is not
the same as the number of \emph{real segments}.

\subsection*{Metric}
\label{sec:metric}

\begin{figure}[htbp]
  \centering \includegraphics{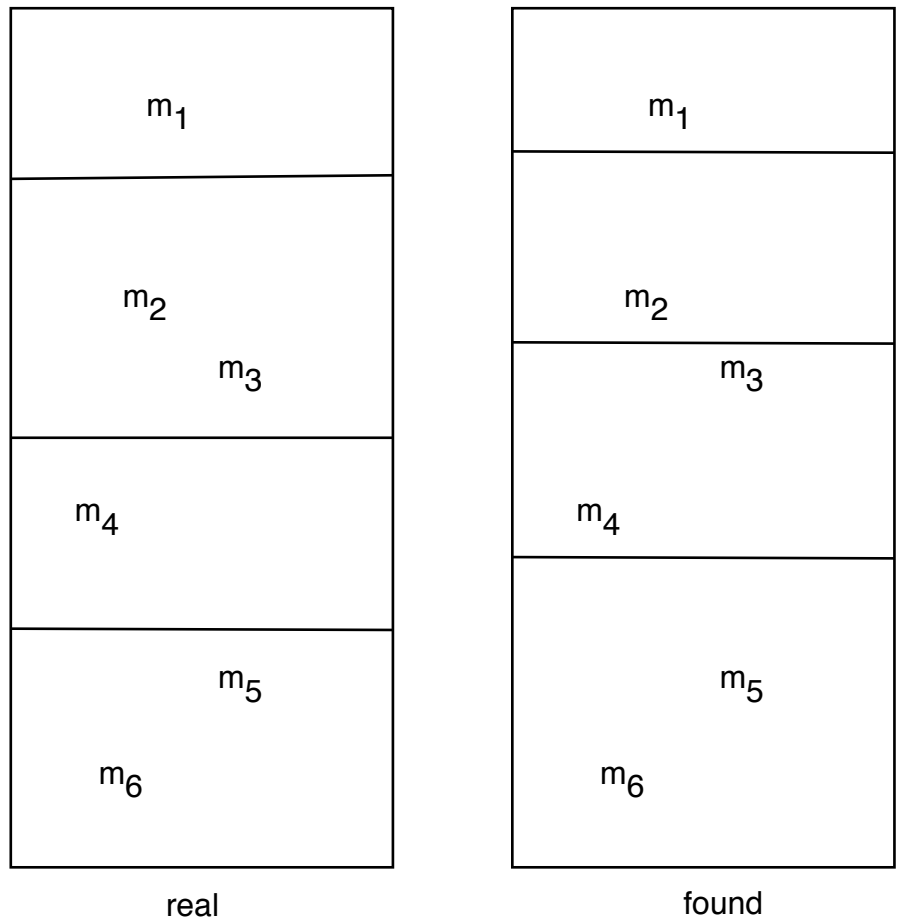}
  \caption{Real vs Found Segmentation}
  \label{fig:rvsf}
\end{figure}

A simple metric giving a pseudo error computation has been set up. For
each word in the document, its position is computed in both
segmentations; for example, in figure \ref{fig:rvsf}, the positions
would be:

\smallskip
\begin{center}

\begin{tabular}{cc}
\begin{math}
  D_r = (1,2,2,3,4,4)
\end{math}
&
\begin{math}
  D_f = (1,2,3,3,4,4)
\end{math}
\\
real positions & found positions
\end{tabular}
\end{center}

Relative position matrices are computed between each pair of words for
both segmentations. The distance
\begin{math}
  M_{i,j} = |D_j - D_i|
\end{math}
is computed for each pair of words i and j where word i occurs before
word j. Triangular matrices illustrated below are obtained.

\smallskip
\begin{center}

\begin{tabular}{cc}
\begin{math}
  R = \left(
\begin {array}{cccccc}
  0&&&&&\\
  1&0&&&&\\
  1&0&0&&&\\
  2&1&1&0&&\\
  3&2&2&1&0&\\
  3&2&2&1&0&0\\
\end {array}
\right)
\end{math}
 &
 \begin{math}
   F = \left(
 \begin {array}{cccccc}
   0&&&&&\\
   1&0&&&&\\
   2&1&0&&&\\
   2&1&0&0&&\\
   3&2&1&1&0&\\
   3&2&1&1&0&0\\
 \end {array}
\right)
 \end{math}
\\
real distance & found distance
 \end{tabular}
\end{center}

\smallskip The error matrix can then be computed as the difference
between the matrices $R$ and $F$:
 \begin{math}
   \label{eq:1}
   E_{i,j} = |R_{i,j} - F_{i,j}|
 \end{math}
 
 \smallskip
\begin{center}

 \begin{math}
   E = \left(
 \begin {array}{cccccc}
   0&&&&&\\
   0&0&&&&\\
   1&0&0&&&\\
   0&0&1&0&&\\
   0&0&1&0&0&\\
   0&0&1&0&0&0\\
 \end {array}
\right)
 \end{math}
\end{center}

\smallskip The average sum of values given by this matrix provides a
concrete metric to evaluate the segmentation. This value can be seen as
the average error in the placement of a word; if it's high, the
segmentation was incorrect, if it is around 0, the segmentation was
almost correct.

This value depends of the number of \emph{real} and \emph{found}
segments which is linked to the size of the text. If the text is
long, there is more chance that there will be more segments (at least
\emph{found} ones). If the algorithm finds 20 segments and there
are, in reality, only 10 segments, the error value could rise up to
20.
 
\subsection*{Results}
\label{sec:results}

\begin{figure}[htbp]
  \centering \input{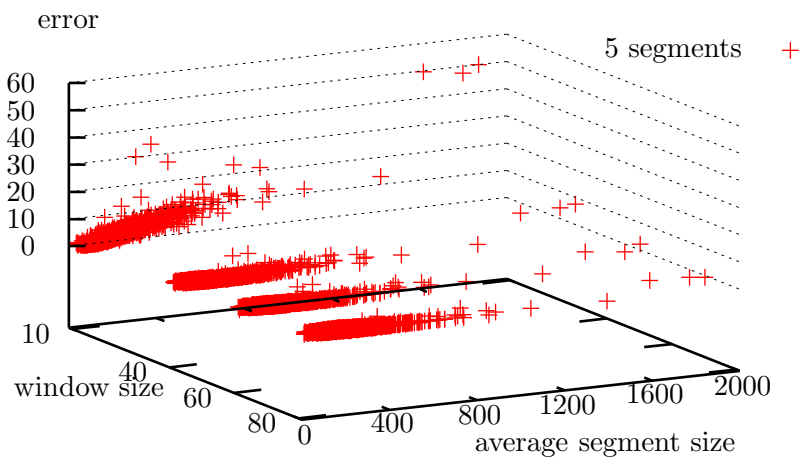}
  \caption[Segmentation of 5 segment texts.]{Segmentation of 5 segment texts with windows of 10, 40, 60
    and 80 words.}
  \label{fig:ev:seg:3d}
\end{figure}

\begin{figure}[htbp]
  \centering \input{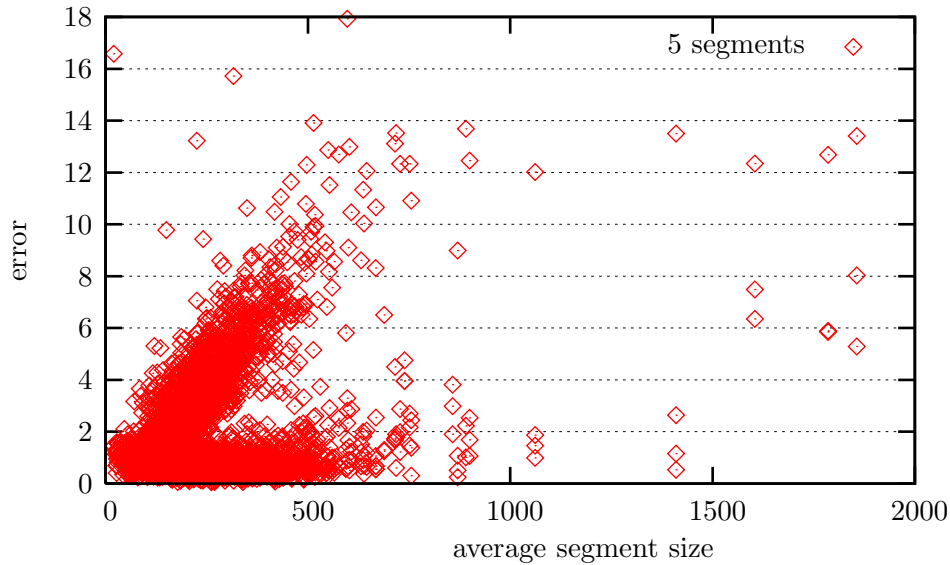}
  \caption[Error vs. Segment Size on 5 segment
    texts.]{Error vs. Segment Size on 5 segment
    texts with windows of 10, 40, 60 and 80 words.}
  \label{fig:ev:seg:34d}
\end{figure}

\begin{figure}[htbp]
  \centering \input{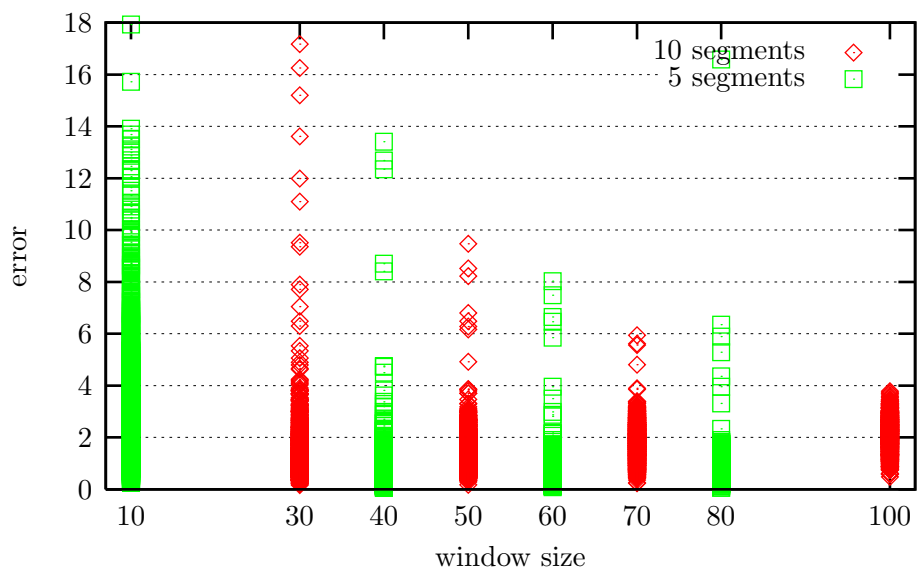}
  \caption{Error vs. Window Size for both segmentations.}
  \label{fig:ev:seg:24}
\end{figure}

The figures \ref{fig:ev:seg:3d}, %
\ref{fig:ev:seg:34d} and \ref{fig:ev:seg:24} give plots of the
evaluation results. The tests were performed on texts having different size
(five and ten segments) and on various window sizes to get a better
view of the variations implied by the algorithm parameters.

Figure \ref{fig:ev:seg:3d} illustrates the differences made
by the window size parameter. This value determines the granularity of
the segmentation, for long text, there should be no need to specify a
low window size as there will not be many more topics than in small
text, but they will be treated in longer segments. For a Window
Size of 10, the error rises quickly even for small text as it's
certain that the algorithm will detect too many segments (on a small
text of 200 words, up to 20 segments could be detected in place of
5).

However, we observe that for windows of size higher than 30, the
difference amongst sizes is not so obvious. In figure
\ref{fig:ev:seg:34d}, we can identify a group of points corresponding
to size 10, but cannot decide in what group of points are the
other ones. Small window size generates too many minima that cannot be
resolved by the smoothing algorithm (see Section \ref{sec:smooth}).

The last figure \ref{fig:ev:seg:24} is the most interesting as it
clearly shows the robustness of the algorithm with higher window size.
It seems that the results for a smaller number of segment are
better for windows of size 60 and 80. This could be explained by the
fact that on 5 segments there are less chances that two consecutive
segments have a similar topic, which will affect the algorithm performance. 

\subsection*{interpretation}
\label{sec:interpretation}

The obtained results enforce the intuitive thought that a small window
size returns bad results. The introduced flow model relies on the fact
that the consecutive windows keep a lexical cohesion as long as the
topic is not changed. The lexical cohesion computation is made on the
criterion of term repetitions; this computation will be wrong for
small windows. For example, a window of 10 words barely contains a
sentence, and the term repetitions between two consecutive sentences
is rarely high as most of the grammatical components change. On
larger windows, repetitions are more likely to occur between two
consecutive windows and give more flexible variation of the windows'
similarity.

Low errors are hard to interpret as it is not obvious how the
error is produced, how to know if it corresponds to many misplaced words with a
low distance error or a few with a high error. Thus it is not evident
to tell what could be done to avoid these small errors. However, there
is a high probability -- because of the actual implementation --
that end or beginning words of certain segments will be badly placed
as the window segmentation does not match the sentence segmentation
provided by the author.

Segmentation could be slightly improved by refining the window
segmentation with criteria like sentence detection. On the other hand,
the poorest results mainly come from an inappropriate window size
selection which cannot be decided without a basic knowledge of the
real segments. This parameter tuning must then be done with the user
input. However, choosing a high window size gives a certain robustness
and could be sufficient for the approximative segmentation used as
preprocessing for topic extraction.

\section{Topic Extraction}
\label{sec:eval:topic-extraction}

The INSPEC corpus, providing rich information on its documents
featured topics, seems to be a choice that would give consistent input
on the topic extraction algorithm effectiveness.  Moreover, topic
annotation in the corpus are generic terms and corresponds perfectly
to what is extracted by our algorithm (see the corpus description in
section \ref{sec:inspec-corpus} for more information). Therefore a
fully automatic evaluation scheme has been developed over this corpus.

\subsection*{Evaluation Scheme}
\label{sec:eval:ext-scheme}

The evaluation scheme should provide information on the effectiveness
of the returned set of concepts compared to the reference annotation
in the corpus. But, there is no way to match each concept in the
produced set with a concept in the reference set because there is a
lot of sibling concepts in our hierarchy that can be interpreted by
the user -- the annotator in our case -- in the same way. For example,
if the annotator chose ``animal'' as an annotation for an entry, but
the extraction algorithm returned ``vertebrate'', there is no big loss
in the semantic of the text. However, no automatic algorithm can match
these concepts together as it has no knowledge of their meaning.

Table \ref{tab:exact-match} presents the standard precision and recall
scores obtained with an evaluation that uses the exact match between
two concepts --i.e. if the concepts are the same, then a match is
found. It's obvious that such evaluation cannot be used in measuring
the performances of our topic extraction system.

\begin{table}[htb]
  \centering
  \begin{tabular}{|c|c|c|c|}
\hline
parameter a&Precision&Recall&F-Measure\\
\hline
0.1&0.0535712&0.0257309&0.030076\\
0.2&0.0535712&0.0257309&0.030076\\
0.3&0.0535712&0.0257309&0.030076\\
0.4&0.0535712&0.0256917&0.0300643\\
0.5&0.0535712&0.0257033&0.030071\\
0.6&0.0535712&0.0256987&0.0300641\\
0.7&0.0540205&0.00890781&0.0145083\\
0.8&0.061942&0.00823537&0.0136329\\
0.9&0.0711698&0.01397&0.0209726\\
\hline
  \end{tabular}
  \caption{Score with exact matching}
  \label{tab:exact-match}
\end{table}

Automatic evaluation requires a metric to compare the extracted
concepts and the provided topics. Similarity metrics in a concept
hierarchy has been used for different purposes in the language
processing field and could be performed by a simple edge counting
scheme in the EDR hierarchy.  \cite{budanitsky01semantic} discusses
different metrics that could be used. The Leacock-Chorodow similarity
measure \cite{leacock98similarity} has been chosen for this evaluation.

This similarity measure is a scaled path length between two concepts.
The scaling is done according to the depth of the conceptual hierarchy as a
whole. This value is defined by:
$$\mathrm{S}(c_i,C_k) = -\ln\left(\frac{\mathrm{d}(c_i,C_k)}{2D}\right)$$
Where
$\mathrm{d}(c_i,C_k)$ is the smallest number of nodes between two
concepts and $D$ is the maximal depth of the hierarchy, which is 17
for the EDR dictionary.

The normalized version of this value, noted $\mathrm{p}(c_i|C_k)$ can
be interpreted as the probability that the concepts $c_i$ and $C_k$
can be \emph{matched}. The evaluation should give an idea of how much
the produced set of concept $\mathrm{Prod} = \{c_1, c_2, \ldots,
c_n\}$ is good enough to describe the set of reference $\mathrm{Ref} =
\{C_1, C_2, \ldots, C_N\}$.

For each concept $c_i$ the probability that it is \emph{correct}
compared to the reference set is the probability that there exists at
least one concept in the reference that can be \emph{matched} with
$c_i$. This is the probability that the event ``no concept in the
reference is \emph{matched} with $c_i$'' does not happen:
$$\mathrm{p}(c_i) = 1 - \prod_{k=1}^{N}(1 - \mathrm{p}(c_i|C_k))$$
In the same way, the probability that at least one concept in $Prod$
is matched to the concept $C_k$ is:
$$\mathrm{p}(C_k) = 1 - \prod_{i=1}^{n}(1 - \mathrm{p}(c_i|C_k))$$

For each concept in the produced set, the first probability can be
used to compute the quality of the extraction. For example, the
probability of \emph{correctness} of an extraction can be computed to
determine its accuracy:
$$\mathrm{A(Prod,Ref)} = \prod_{k=1}^{n} \mathrm{p}(c_i)$$

However, this value is not always meaningful. For example, if some
concepts in $\mathrm{Prod}$ are \emph{wrong} -- i.e. they do not
correspond to any concept in $\mathrm{Ref}$ -- the score will be low.
In the same way, if some concepts in $\mathrm{Ref}$ are not
represented by a concept in $\mathrm{Prod}$, the score will be
low and there will be no way to determine for which of
these two reasons the score is \emph{bad}.

In Natural Language Processing, \gloss[word]{precision} and
\gloss[word]{recall} are often used to describe the effectiveness of
an extraction algorithm. The precision is the ratio between the number
of relevant topics extracted and the total number of topics retrieved,
whereas the recall is the ratio of relevant topics extracted over the
total number of reference topics known.

However, in our evaluation, there is no way to know which concept in
$\mathrm{Prod}$ is to be matched to a concept in $\mathrm{Ref}$, but
we can still describe these score. If we interpret:
\begin{itemize}
\item[$\mathrm{p}(c_i)$] as the probability that the produced concept $c_i$
  matches at least one reference concept,
\item[$\mathrm{p}(C_k)$] as the probability that a reference concept
  $C_k$ is matched by at least one concept produced.
\end{itemize}
Then the expectation of each score can be computed. Let:
\begin{itemize}
\item[$d_i$] be the function that is equal to 1 when the produced concept
  $c_i$ is matched and 0 otherwise,
\item[$D_k$] be the function that is equal to 1 when the reference
  concept $C_k$ is matched and 0 otherwise.
\end{itemize}
The scores can be written:
\begin{align*}
\mathrm{P(Prod,Ref)} & = E(P) =
 E\left(\sum_{i=1}^{n}\left(\frac{d_i}{n}\right)\right)\\
 & = \sum_{i=1}^{n}\left(\frac{E(d_i)}{n}\right)\\
& = \frac{1}{n} \times \sum_{i=1}^{n}\mathrm{p}(c_i)
\end{align*}
and:
\begin{align*}
\mathrm{R(Prod,Ref)} & = E(R) \\
& = E\left(\sum_{k=1}^{N}\left(\frac{D_k}{N}\right)\right)\\
& = \frac{1}{N} \times \sum_{i=k}^{N}\mathrm{p}(C_k)
\end{align*}
The usual parametric F-measure can also be derived from these scores:
$$\mathrm{F(Prod,Ref)} = \frac{(b^2+1).\mathrm{P(Prod,Ref)}.\mathrm{R(Prod,Ref)}}{b^2.\mathrm{P(Prod,Ref)}+\mathrm{R(Prod,Ref)}}$$

Lets look at what is expected for these value. A good algorithm should
reach a value of $\mathrm{R(Prod,Ref)} = 1$, which means that
$\frac{1}{N} \times \sum_{k=1}^{N}\mathrm{p}(C_k)$ is equal to 1, or
$\sum_{k=1}^{N}\mathrm{p}(C_k) = N$. Therefore:
\begin{align*}
\forall C_k \vdash & \prod_{i=1}^{n}(1 - \mathrm{p}(c_i|C_k)) = 0 \\
\forall C_k \exists c_i \vdash & 1 - \mathrm{p}(c_i|C_k) = 0 \\
\forall C_k \exists c_i \vdash & \mathrm{p}(c_i|C_k) = 1 \\
\end{align*}
That means that for all concepts $C_k$ in the reference set $\mathrm{Ref}$,
there exists at least one concept $c_i$ in the produced set $\mathrm{Prod}$
that is \emph{equivalent} to $C_k$. For small values of $a$, the
extracted concepts are generic and there is a small chance that each
concept in the reference is perfectly \emph{matched} by a concept in
the production.

In the same way, $\mathrm{P(Prod,Ref)} = 1$ means that:
$$\forall c_i \exists C_k \vdash \mathrm{p}(c_i|C_k) = 1$$
All the concepts $c_i$ in the production perfectly \emph{match} at
least one concept $C_k$ in the reference. For big values of $a$, the
set of concepts extracted contains more specific concepts. It should
also contains a bigger number of concepts, therefore, the chances this
condition is verified are small --i.e. there is too much noise.

\subsection*{Parameter Optimization}
\label{sec:param-optim}

As explained in section \ref{sec:score-combination}, the algorithm is
controlled by a parameter $a$ that is supposed to regulate the number
of concepts extracted -- by controlling the level of the cut in the
hierarchy. The corpus that has been created for the evaluation does
not always offer the same amount of concepts in each entry annotation.
The first experiment to optimize, for each entry, which value of $a$
is used for the evaluation.

\begin{figure}[htbp]
  \centering \setlength{\unitlength}{0.240900pt}
\ifx\plotpoint\undefined\newsavebox{\plotpoint}\fi
\sbox{\plotpoint}{\rule[-0.200pt]{0.400pt}{0.400pt}}%
\begin{picture}(1500,900)(0,0)
\sbox{\plotpoint}{\rule[-0.200pt]{0.400pt}{0.400pt}}%
\put(160.0,123.0){\rule[-0.200pt]{4.818pt}{0.400pt}}
\put(140,123){\makebox(0,0)[r]{ 0.6}}
\put(1419.0,123.0){\rule[-0.200pt]{4.818pt}{0.400pt}}
\put(160.0,215.0){\rule[-0.200pt]{4.818pt}{0.400pt}}
\put(140,215){\makebox(0,0)[r]{ 0.65}}
\put(1419.0,215.0){\rule[-0.200pt]{4.818pt}{0.400pt}}
\put(160.0,307.0){\rule[-0.200pt]{4.818pt}{0.400pt}}
\put(140,307){\makebox(0,0)[r]{ 0.7}}
\put(1419.0,307.0){\rule[-0.200pt]{4.818pt}{0.400pt}}
\put(160.0,399.0){\rule[-0.200pt]{4.818pt}{0.400pt}}
\put(140,399){\makebox(0,0)[r]{ 0.75}}
\put(1419.0,399.0){\rule[-0.200pt]{4.818pt}{0.400pt}}
\put(160.0,492.0){\rule[-0.200pt]{4.818pt}{0.400pt}}
\put(140,492){\makebox(0,0)[r]{ 0.8}}
\put(1419.0,492.0){\rule[-0.200pt]{4.818pt}{0.400pt}}
\put(160.0,584.0){\rule[-0.200pt]{4.818pt}{0.400pt}}
\put(140,584){\makebox(0,0)[r]{ 0.85}}
\put(1419.0,584.0){\rule[-0.200pt]{4.818pt}{0.400pt}}
\put(160.0,676.0){\rule[-0.200pt]{4.818pt}{0.400pt}}
\put(140,676){\makebox(0,0)[r]{ 0.9}}
\put(1419.0,676.0){\rule[-0.200pt]{4.818pt}{0.400pt}}
\put(160.0,768.0){\rule[-0.200pt]{4.818pt}{0.400pt}}
\put(140,768){\makebox(0,0)[r]{ 0.95}}
\put(1419.0,768.0){\rule[-0.200pt]{4.818pt}{0.400pt}}
\put(160.0,860.0){\rule[-0.200pt]{4.818pt}{0.400pt}}
\put(140,860){\makebox(0,0)[r]{ 1}}
\put(1419.0,860.0){\rule[-0.200pt]{4.818pt}{0.400pt}}
\put(160.0,123.0){\rule[-0.200pt]{0.400pt}{4.818pt}}
\put(160,82){\makebox(0,0){ 0.1}}
\put(160.0,840.0){\rule[-0.200pt]{0.400pt}{4.818pt}}
\put(320.0,123.0){\rule[-0.200pt]{0.400pt}{4.818pt}}
\put(320,82){\makebox(0,0){ 0.2}}
\put(320.0,840.0){\rule[-0.200pt]{0.400pt}{4.818pt}}
\put(480.0,123.0){\rule[-0.200pt]{0.400pt}{4.818pt}}
\put(480,82){\makebox(0,0){ 0.3}}
\put(480.0,840.0){\rule[-0.200pt]{0.400pt}{4.818pt}}
\put(640.0,123.0){\rule[-0.200pt]{0.400pt}{4.818pt}}
\put(640,82){\makebox(0,0){ 0.4}}
\put(640.0,840.0){\rule[-0.200pt]{0.400pt}{4.818pt}}
\put(800.0,123.0){\rule[-0.200pt]{0.400pt}{4.818pt}}
\put(800,82){\makebox(0,0){ 0.5}}
\put(800.0,840.0){\rule[-0.200pt]{0.400pt}{4.818pt}}
\put(959.0,123.0){\rule[-0.200pt]{0.400pt}{4.818pt}}
\put(959,82){\makebox(0,0){ 0.6}}
\put(959.0,840.0){\rule[-0.200pt]{0.400pt}{4.818pt}}
\put(1119.0,123.0){\rule[-0.200pt]{0.400pt}{4.818pt}}
\put(1119,82){\makebox(0,0){ 0.7}}
\put(1119.0,840.0){\rule[-0.200pt]{0.400pt}{4.818pt}}
\put(1279.0,123.0){\rule[-0.200pt]{0.400pt}{4.818pt}}
\put(1279,82){\makebox(0,0){ 0.8}}
\put(1279.0,840.0){\rule[-0.200pt]{0.400pt}{4.818pt}}
\put(1439.0,123.0){\rule[-0.200pt]{0.400pt}{4.818pt}}
\put(1439,82){\makebox(0,0){ 0.9}}
\put(1439.0,840.0){\rule[-0.200pt]{0.400pt}{4.818pt}}
\put(160.0,123.0){\rule[-0.200pt]{308.111pt}{0.400pt}}
\put(1439.0,123.0){\rule[-0.200pt]{0.400pt}{177.543pt}}
\put(160.0,860.0){\rule[-0.200pt]{308.111pt}{0.400pt}}
\put(160.0,123.0){\rule[-0.200pt]{0.400pt}{177.543pt}}
\put(799,21){\makebox(0,0){a}}
\put(1259,584){\makebox(0,0)[r]{Accuracy}}
\put(1279.0,584.0){\rule[-0.200pt]{24.090pt}{0.400pt}}
\put(160,744){\usebox{\plotpoint}}
\put(480,743.67){\rule{38.544pt}{0.400pt}}
\multiput(480.00,743.17)(80.000,1.000){2}{\rule{19.272pt}{0.400pt}}
\put(160.0,744.0){\rule[-0.200pt]{77.088pt}{0.400pt}}
\multiput(959.00,745.59)(17.742,0.477){7}{\rule{12.900pt}{0.115pt}}
\multiput(959.00,744.17)(133.225,5.000){2}{\rule{6.450pt}{0.400pt}}
\multiput(1119.00,750.58)(5.106,0.494){29}{\rule{4.100pt}{0.119pt}}
\multiput(1119.00,749.17)(151.490,16.000){2}{\rule{2.050pt}{0.400pt}}
\multiput(1279.00,766.58)(3.867,0.496){39}{\rule{3.148pt}{0.119pt}}
\multiput(1279.00,765.17)(153.467,21.000){2}{\rule{1.574pt}{0.400pt}}
\put(640.0,745.0){\rule[-0.200pt]{76.847pt}{0.400pt}}
\put(1259,543){\makebox(0,0)[r]{Precision}}
\multiput(1279,543)(20.756,0.000){5}{\usebox{\plotpoint}}
\put(1379,543){\usebox{\plotpoint}}
\put(160,268){\usebox{\plotpoint}}
\multiput(160,268)(20.756,0.000){8}{\usebox{\plotpoint}}
\multiput(320,268)(20.756,0.000){8}{\usebox{\plotpoint}}
\multiput(480,268)(20.756,0.000){8}{\usebox{\plotpoint}}
\multiput(640,268)(20.756,0.000){7}{\usebox{\plotpoint}}
\multiput(800,268)(20.754,-0.261){8}{\usebox{\plotpoint}}
\multiput(959,266)(20.707,-1.424){8}{\usebox{\plotpoint}}
\multiput(1119,255)(20.377,-3.948){8}{\usebox{\plotpoint}}
\multiput(1279,224)(19.102,-8.118){8}{\usebox{\plotpoint}}
\put(1439,156){\usebox{\plotpoint}}
\sbox{\plotpoint}{\rule[-0.400pt]{0.800pt}{0.800pt}}%
\sbox{\plotpoint}{\rule[-0.200pt]{0.400pt}{0.400pt}}%
\put(1259,502){\makebox(0,0)[r]{Recall}}
\sbox{\plotpoint}{\rule[-0.400pt]{0.800pt}{0.800pt}}%
\put(1279.0,502.0){\rule[-0.400pt]{24.090pt}{0.800pt}}
\put(160,676){\usebox{\plotpoint}}
\put(480,674.84){\rule{38.544pt}{0.800pt}}
\multiput(480.00,674.34)(80.000,1.000){2}{\rule{19.272pt}{0.800pt}}
\put(160.0,676.0){\rule[-0.400pt]{77.088pt}{0.800pt}}
\put(800,675.84){\rule{38.303pt}{0.800pt}}
\multiput(800.00,675.34)(79.500,1.000){2}{\rule{19.152pt}{0.800pt}}
\multiput(959.00,679.41)(7.181,0.511){17}{\rule{10.867pt}{0.123pt}}
\multiput(959.00,676.34)(137.446,12.000){2}{\rule{5.433pt}{0.800pt}}
\multiput(1119.00,691.41)(2.194,0.503){67}{\rule{3.659pt}{0.121pt}}
\multiput(1119.00,688.34)(152.405,37.000){2}{\rule{1.830pt}{0.800pt}}
\multiput(1279.00,728.41)(1.582,0.502){95}{\rule{2.710pt}{0.121pt}}
\multiput(1279.00,725.34)(154.376,51.000){2}{\rule{1.355pt}{0.800pt}}
\put(640.0,677.0){\rule[-0.400pt]{38.544pt}{0.800pt}}
\sbox{\plotpoint}{\rule[-0.500pt]{1.000pt}{1.000pt}}%
\sbox{\plotpoint}{\rule[-0.200pt]{0.400pt}{0.400pt}}%
\put(1259,461){\makebox(0,0)[r]{F-Measure}}
\sbox{\plotpoint}{\rule[-0.500pt]{1.000pt}{1.000pt}}%
\multiput(1279,461)(20.756,0.000){5}{\usebox{\plotpoint}}
\put(1379,461){\usebox{\plotpoint}}
\put(160,395){\usebox{\plotpoint}}
\multiput(160,395)(20.756,0.000){8}{\usebox{\plotpoint}}
\multiput(320,395)(20.756,0.000){8}{\usebox{\plotpoint}}
\multiput(480,395)(20.756,0.000){8}{\usebox{\plotpoint}}
\multiput(640,395)(20.756,0.000){7}{\usebox{\plotpoint}}
\multiput(800,395)(20.756,0.000){8}{\usebox{\plotpoint}}
\multiput(959,395)(20.736,-0.907){8}{\usebox{\plotpoint}}
\multiput(1119,388)(20.625,-2.320){7}{\usebox{\plotpoint}}
\multiput(1279,370)(19.980,-5.619){8}{\usebox{\plotpoint}}
\put(1439,325){\usebox{\plotpoint}}
\sbox{\plotpoint}{\rule[-0.200pt]{0.400pt}{0.400pt}}%
\put(160.0,123.0){\rule[-0.200pt]{308.111pt}{0.400pt}}
\put(1439.0,123.0){\rule[-0.200pt]{0.400pt}{177.543pt}}
\put(160.0,860.0){\rule[-0.200pt]{308.111pt}{0.400pt}}
\put(160.0,123.0){\rule[-0.200pt]{0.400pt}{177.543pt}}
\end{picture}
  \caption{Comparison of the algorithm results with variation of a}
  \label{fig:a-param}
\end{figure}
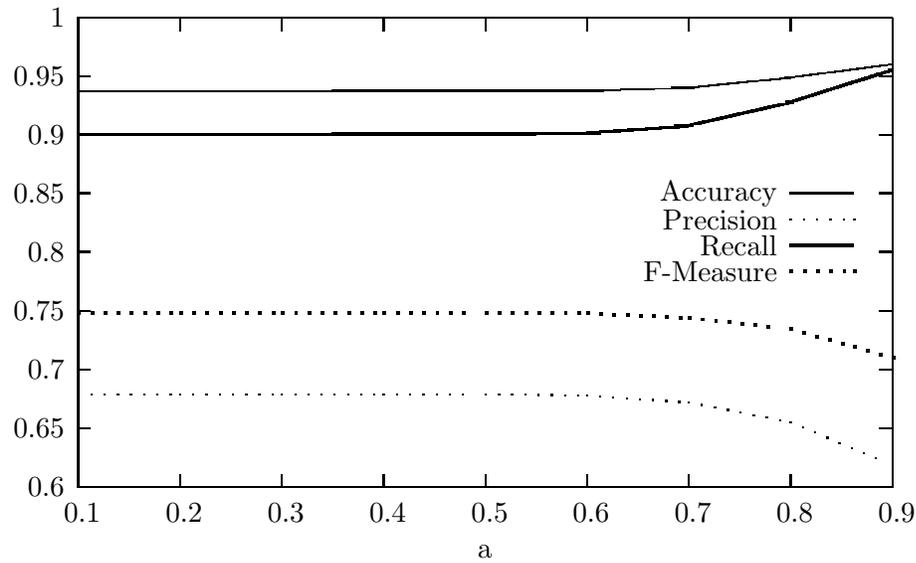

In our system, the smallest number of extracted concepts is, the more
generic they are. As shown in figure \ref{fig:a-param}, nine
evaluations have been done on every entry with a different value for
the parameter $a$. This shows the evolution of the algorithm
\emph{effectiveness}. 

Observing the average results obtained for each value of $a$, it is
obvious that $a$ has a really small impact on the algorithm
performances. The F-measure is quasi constant until $a=0.6$. Then, the
Recall rise and the Precision falls because the algorithm outputs a
bigger number of concepts that will be more precise -- i.e.  they
match more concepts in $\mathrm{Ref}$ but produce more noise.

For value of $a$ under 0.6, the evaluation metric gives comparable
results for the similarity of each output with the reference. Which
means that the semantic similarity between each set of extracted
concept is small. This could mean two things:
\begin{enumerate}
\item the extracted cuts are not very distant in the hierarchy,
\item the similarity metric used considers that the cuts are similar.
\end{enumerate}

\begin{figure}[!p]
  \centering \includegraphics[width=\textwidth,height=\textheight]{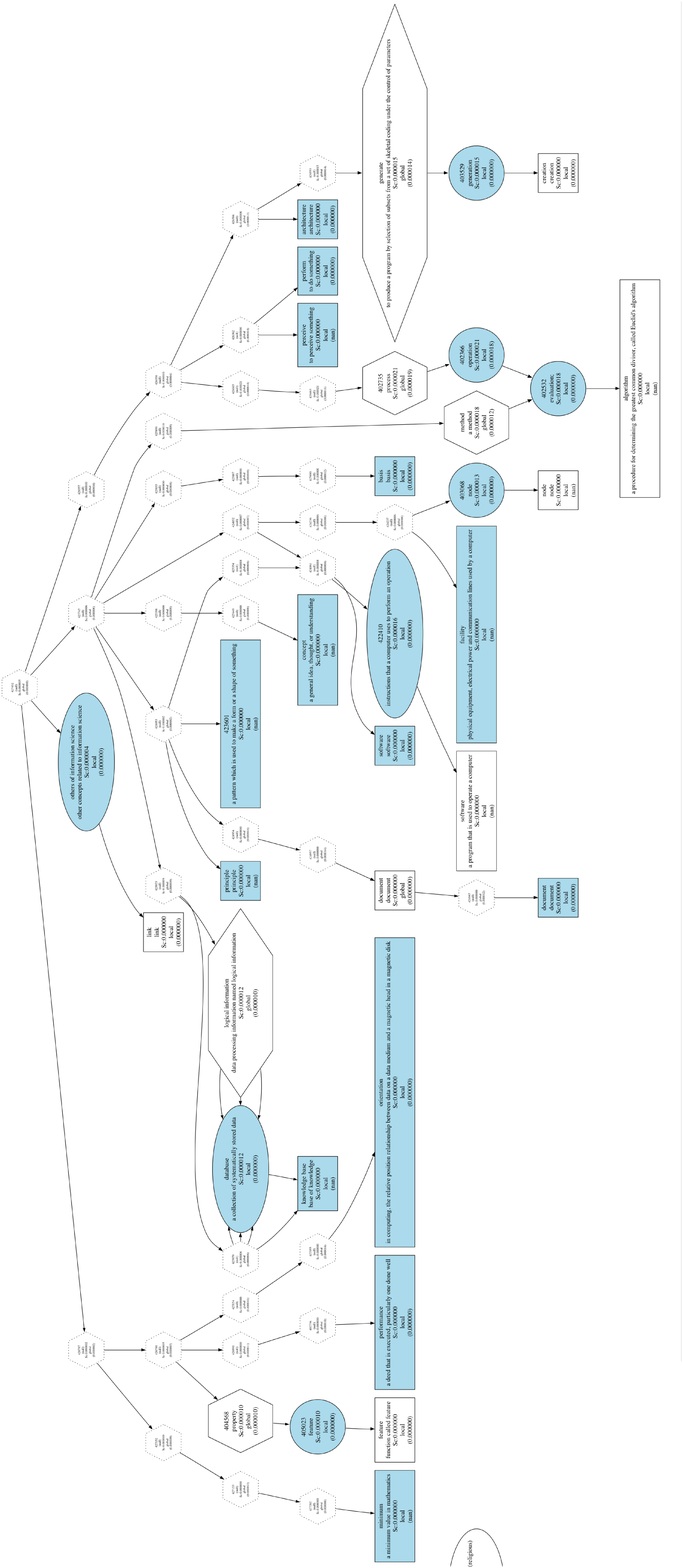}
  \caption{Excerpt of an extracted hierarchy}
  \label{fig:badGraph}
\end{figure}

Both causes can be explained by looking at the hierarchy extracted
during an algorithm run. Such hierarchy is too big to be included in
the document, but figure \ref{fig:badGraph} displays an example part
of one hierarchy. Colored nodes are the ones that are selected in the
cut, rectangles are leaves and hexagons are nodes that should be
extended (see algorithm \ref{alg:sec:cut-extraction}). The little
dashed nodes are \emph{anonymous} concepts (see section
\ref{sec:edr:anonymous-concepts}) that are ignored in the algorithm
processing (an anonymous concepts has no real meaning).

There is many \emph{anonymous} concepts in the hierarchy and if only
the non-anonymous ones are considered, the hierarchy is quite flat.
The definition of the semantic granularity is poor so the algorithm is
unable to select a lot of cut between the leaves and the root. In the
same way, there won't be a lot of nodes counted in the similarity
metric and many concepts are considered to be \emph{similar}.

With such results, it's hard to choose a best $a$ value for each
document to perform the evaluation. Looking at the plot, it would be
better to choose a value between $a=0.1$ and $a=0.7$ where the results
are the bests. However, the plot also displays that the extraction
returned \emph{equivalent} results between these values, which means
that the extracted concepts are generic ones. We believe that studying
the results obtained with $a$ between 0.6 and 0.9 is more interesting
as it will display the behaviour of the extraction algorithm when it
extract \emph{specific} concepts that might be \emph{nearer} to the
reference annotation.

\subsection*{Results Interpretation}
\label{sec:results-interpr}

A standard evaluation procedure has been adapted to compute the
average precision and recall:
\begin{enumerate}
\item all the probabilities $p(c_i)$ and $p(C_k)$ is computed for each
  documents in the evaluation corpus,
\item the concepts $c_i$ in $\mathrm{Prod}$ and $C_k$ in
  $\mathrm{Ref}$ is sorted by descending probabilities,
\item for each value of the threshold $\Theta$ in [0,1[, a
  Precision/Recall pair is computed taking into account only the
  concepts for which $p(c_i) > \Theta$ or $p(C_i) > \Theta$,
\item the average Precision, Recall and F-Measure is computed for each
  value of $\Theta$.
\end{enumerate}

For example, for a document with the sets of concepts given in table
\ref{tab:prob-ex}, and $\Theta=0.97$, the extracted concepts
$\{393453,358943,359232,358845\}$ and the produced concepts
$\{22607,22606\}$ are considered. And the scores are:
$$\mathrm{P(Pord,Ref)} = \frac{3.99999629}{4}$$
$$\mathrm{P(Pord,Ref)} = \frac{1.96710571}{8}$$

\begin{table}[h]
\begin{tabular}{|l|r||l|r|}
\multicolumn{2}{c}{$Prod$} &\multicolumn{2}{c}{$Ref$}\\ 
\hline
$c_i$&$p(c_i)$&$C_k$&$p(C_k)$\\
\hline
\underline{393453}&0.99999954&\underline{22607} &0.98911961\\
\underline{358943}&0.99999950&\underline{22606} &0.97798610\\
\underline{359232}&0.99999866&84062 &0.74886709\\
\underline{358845}&0.99999859&395795&0.74886709\\
460912&0.94022170&391576&0.74886709\\
404568&0.91138175&364931&0.71981943\\
402831&0.90216440&393495&0.71068739\\
423743&0.88603944&364267&0.71068739\\
422997&0.88603944&&\\
\hline
\end{tabular}
\centering
  \caption{Example of the probability obtained}
  \label{tab:prob-ex}
\end{table}

\begin{figure}[htb]
  \centering \setlength{\unitlength}{0.240900pt}
\ifx\plotpoint\undefined\newsavebox{\plotpoint}\fi
\begin{picture}(1500,900)(0,0)
\sbox{\plotpoint}{\rule[-0.200pt]{0.400pt}{0.400pt}}%
\put(201.0,123.0){\rule[-0.200pt]{4.818pt}{0.400pt}}
\put(181,123){\makebox(0,0)[r]{ 0.81}}
\put(1419.0,123.0){\rule[-0.200pt]{4.818pt}{0.400pt}}
\put(201.0,205.0){\rule[-0.200pt]{4.818pt}{0.400pt}}
\put(181,205){\makebox(0,0)[r]{ 0.82}}
\put(1419.0,205.0){\rule[-0.200pt]{4.818pt}{0.400pt}}
\put(201.0,287.0){\rule[-0.200pt]{4.818pt}{0.400pt}}
\put(181,287){\makebox(0,0)[r]{ 0.83}}
\put(1419.0,287.0){\rule[-0.200pt]{4.818pt}{0.400pt}}
\put(201.0,369.0){\rule[-0.200pt]{4.818pt}{0.400pt}}
\put(181,369){\makebox(0,0)[r]{ 0.84}}
\put(1419.0,369.0){\rule[-0.200pt]{4.818pt}{0.400pt}}
\put(201.0,451.0){\rule[-0.200pt]{4.818pt}{0.400pt}}
\put(181,451){\makebox(0,0)[r]{ 0.85}}
\put(1419.0,451.0){\rule[-0.200pt]{4.818pt}{0.400pt}}
\put(201.0,532.0){\rule[-0.200pt]{4.818pt}{0.400pt}}
\put(181,532){\makebox(0,0)[r]{ 0.86}}
\put(1419.0,532.0){\rule[-0.200pt]{4.818pt}{0.400pt}}
\put(201.0,614.0){\rule[-0.200pt]{4.818pt}{0.400pt}}
\put(181,614){\makebox(0,0)[r]{ 0.87}}
\put(1419.0,614.0){\rule[-0.200pt]{4.818pt}{0.400pt}}
\put(201.0,696.0){\rule[-0.200pt]{4.818pt}{0.400pt}}
\put(181,696){\makebox(0,0)[r]{ 0.88}}
\put(1419.0,696.0){\rule[-0.200pt]{4.818pt}{0.400pt}}
\put(201.0,778.0){\rule[-0.200pt]{4.818pt}{0.400pt}}
\put(181,778){\makebox(0,0)[r]{ 0.89}}
\put(1419.0,778.0){\rule[-0.200pt]{4.818pt}{0.400pt}}
\put(201.0,860.0){\rule[-0.200pt]{4.818pt}{0.400pt}}
\put(181,860){\makebox(0,0)[r]{ 0.9}}
\put(1419.0,860.0){\rule[-0.200pt]{4.818pt}{0.400pt}}
\put(201.0,123.0){\rule[-0.200pt]{0.400pt}{4.818pt}}
\put(201,82){\makebox(0,0){ 0.2}}
\put(201.0,840.0){\rule[-0.200pt]{0.400pt}{4.818pt}}
\put(378.0,123.0){\rule[-0.200pt]{0.400pt}{4.818pt}}
\put(378,82){\makebox(0,0){ 0.3}}
\put(378.0,840.0){\rule[-0.200pt]{0.400pt}{4.818pt}}
\put(555.0,123.0){\rule[-0.200pt]{0.400pt}{4.818pt}}
\put(555,82){\makebox(0,0){ 0.4}}
\put(555.0,840.0){\rule[-0.200pt]{0.400pt}{4.818pt}}
\put(732.0,123.0){\rule[-0.200pt]{0.400pt}{4.818pt}}
\put(732,82){\makebox(0,0){ 0.5}}
\put(732.0,840.0){\rule[-0.200pt]{0.400pt}{4.818pt}}
\put(908.0,123.0){\rule[-0.200pt]{0.400pt}{4.818pt}}
\put(908,82){\makebox(0,0){ 0.6}}
\put(908.0,840.0){\rule[-0.200pt]{0.400pt}{4.818pt}}
\put(1085.0,123.0){\rule[-0.200pt]{0.400pt}{4.818pt}}
\put(1085,82){\makebox(0,0){ 0.7}}
\put(1085.0,840.0){\rule[-0.200pt]{0.400pt}{4.818pt}}
\put(1262.0,123.0){\rule[-0.200pt]{0.400pt}{4.818pt}}
\put(1262,82){\makebox(0,0){ 0.8}}
\put(1262.0,840.0){\rule[-0.200pt]{0.400pt}{4.818pt}}
\put(1439.0,123.0){\rule[-0.200pt]{0.400pt}{4.818pt}}
\put(1439,82){\makebox(0,0){ 0.9}}
\put(1439.0,840.0){\rule[-0.200pt]{0.400pt}{4.818pt}}
\put(201.0,123.0){\rule[-0.200pt]{298.234pt}{0.400pt}}
\put(1439.0,123.0){\rule[-0.200pt]{0.400pt}{177.543pt}}
\put(201.0,860.0){\rule[-0.200pt]{298.234pt}{0.400pt}}
\put(201.0,123.0){\rule[-0.200pt]{0.400pt}{177.543pt}}
\put(40,491){\makebox(0,0){\begin{sideways}Precision\end{sideways}}}
\put(820,21){\makebox(0,0){Recall}}
\put(358,778){\makebox(0,0)[r]{a=0.6}}
\put(378.0,778.0){\rule[-0.200pt]{24.090pt}{0.400pt}}
\put(378,224){\usebox{\plotpoint}}
\multiput(378.58,224.00)(0.500,1.128){351}{\rule{0.120pt}{1.002pt}}
\multiput(377.17,224.00)(177.000,396.921){2}{\rule{0.400pt}{0.501pt}}
\multiput(555.00,623.58)(0.784,0.499){223}{\rule{0.727pt}{0.120pt}}
\multiput(555.00,622.17)(175.492,113.000){2}{\rule{0.363pt}{0.400pt}}
\multiput(732.00,736.58)(1.882,0.498){91}{\rule{1.598pt}{0.120pt}}
\multiput(732.00,735.17)(172.684,47.000){2}{\rule{0.799pt}{0.400pt}}
\multiput(908.00,783.58)(6.038,0.494){27}{\rule{4.820pt}{0.119pt}}
\multiput(908.00,782.17)(166.996,15.000){2}{\rule{2.410pt}{0.400pt}}
\multiput(1085.58,795.52)(0.500,-0.621){351}{\rule{0.120pt}{0.597pt}}
\multiput(1084.17,796.76)(177.000,-218.761){2}{\rule{0.400pt}{0.299pt}}
\put(378,224){\raisebox{-.8pt}{\makebox(0,0){$\Diamond$}}}
\put(555,623){\raisebox{-.8pt}{\makebox(0,0){$\Diamond$}}}
\put(732,736){\raisebox{-.8pt}{\makebox(0,0){$\Diamond$}}}
\put(908,783){\raisebox{-.8pt}{\makebox(0,0){$\Diamond$}}}
\put(1085,798){\raisebox{-.8pt}{\makebox(0,0){$\Diamond$}}}
\put(1262,578){\raisebox{-.8pt}{\makebox(0,0){$\Diamond$}}}
\put(428,778){\raisebox{-.8pt}{\makebox(0,0){$\Diamond$}}}
\sbox{\plotpoint}{\rule[-0.600pt]{1.200pt}{1.200pt}}%
\sbox{\plotpoint}{\rule[-0.200pt]{0.400pt}{0.400pt}}%
\put(358,737){\makebox(0,0)[r]{a=0.7}}
\sbox{\plotpoint}{\rule[-0.600pt]{1.200pt}{1.200pt}}%
\put(378.0,737.0){\rule[-0.600pt]{24.090pt}{1.200pt}}
\put(378,195){\usebox{\plotpoint}}
\multiput(380.24,195.00)(0.500,0.841){344}{\rule{0.120pt}{2.320pt}}
\multiput(375.51,195.00)(177.000,293.184){2}{\rule{1.200pt}{1.160pt}}
\multiput(557.24,493.00)(0.500,0.597){344}{\rule{0.120pt}{1.737pt}}
\multiput(552.51,493.00)(177.000,208.394){2}{\rule{1.200pt}{0.869pt}}
\multiput(732.00,707.24)(1.469,0.500){110}{\rule{3.820pt}{0.120pt}}
\multiput(732.00,702.51)(168.071,60.000){2}{\rule{1.910pt}{1.200pt}}
\multiput(908.00,767.24)(3.926,0.501){36}{\rule{9.535pt}{0.121pt}}
\multiput(908.00,762.51)(157.210,23.000){2}{\rule{4.767pt}{1.200pt}}
\multiput(1087.24,780.65)(0.500,-0.611){344}{\rule{0.120pt}{1.771pt}}
\multiput(1082.51,784.32)(177.000,-213.324){2}{\rule{1.200pt}{0.886pt}}
\put(378,195){\makebox(0,0){$\triangle$}}
\put(555,493){\makebox(0,0){$\triangle$}}
\put(732,705){\makebox(0,0){$\triangle$}}
\put(908,765){\makebox(0,0){$\triangle$}}
\put(1085,788){\makebox(0,0){$\triangle$}}
\put(1262,571){\makebox(0,0){$\triangle$}}
\put(428,737){\makebox(0,0){$\triangle$}}
\sbox{\plotpoint}{\rule[-0.400pt]{0.800pt}{0.800pt}}%
\sbox{\plotpoint}{\rule[-0.200pt]{0.400pt}{0.400pt}}%
\put(358,696){\makebox(0,0)[r]{a=0.8}}
\sbox{\plotpoint}{\rule[-0.400pt]{0.800pt}{0.800pt}}%
\put(378.0,696.0){\rule[-0.400pt]{24.090pt}{0.800pt}}
\put(555,193){\usebox{\plotpoint}}
\multiput(556.41,193.00)(0.501,0.715){347}{\rule{0.121pt}{1.344pt}}
\multiput(553.34,193.00)(177.000,250.211){2}{\rule{0.800pt}{0.672pt}}
\multiput(733.41,446.00)(0.501,0.591){345}{\rule{0.121pt}{1.145pt}}
\multiput(730.34,446.00)(176.000,205.623){2}{\rule{0.800pt}{0.573pt}}
\multiput(908.00,655.41)(0.813,0.501){211}{\rule{1.499pt}{0.121pt}}
\multiput(908.00,652.34)(173.889,109.000){2}{\rule{0.750pt}{0.800pt}}
\multiput(1085.00,761.09)(1.824,-0.502){91}{\rule{3.090pt}{0.121pt}}
\multiput(1085.00,761.34)(170.587,-49.000){2}{\rule{1.545pt}{0.800pt}}
\multiput(1262.00,712.09)(0.720,-0.501){239}{\rule{1.351pt}{0.121pt}}
\multiput(1262.00,712.34)(174.195,-123.000){2}{\rule{0.676pt}{0.800pt}}
\put(555,193){\raisebox{-.8pt}{\makebox(0,0){$\Box$}}}
\put(732,446){\raisebox{-.8pt}{\makebox(0,0){$\Box$}}}
\put(908,654){\raisebox{-.8pt}{\makebox(0,0){$\Box$}}}
\put(1085,763){\raisebox{-.8pt}{\makebox(0,0){$\Box$}}}
\put(1262,714){\raisebox{-.8pt}{\makebox(0,0){$\Box$}}}
\put(1439,591){\raisebox{-.8pt}{\makebox(0,0){$\Box$}}}
\put(428,696){\raisebox{-.8pt}{\makebox(0,0){$\Box$}}}
\sbox{\plotpoint}{\rule[-0.500pt]{1.000pt}{1.000pt}}%
\sbox{\plotpoint}{\rule[-0.200pt]{0.400pt}{0.400pt}}%
\put(358,655){\makebox(0,0)[r]{a=0.9}}
\sbox{\plotpoint}{\rule[-0.500pt]{1.000pt}{1.000pt}}%
\multiput(378,655)(20.756,0.000){5}{\usebox{\plotpoint}}
\put(478,655){\usebox{\plotpoint}}
\put(908,293){\usebox{\plotpoint}}
\multiput(908,293)(14.349,14.997){13}{\usebox{\plotpoint}}
\multiput(1085,478)(10.786,17.733){16}{\usebox{\plotpoint}}
\multiput(1262,769)(14.228,-15.112){13}{\usebox{\plotpoint}}
\put(1439,581){\usebox{\plotpoint}}
\put(908,293){\makebox(0,0){$\times$}}
\put(1085,478){\makebox(0,0){$\times$}}
\put(1262,769){\makebox(0,0){$\times$}}
\put(1439,581){\makebox(0,0){$\times$}}
\put(428,655){\makebox(0,0){$\times$}}
\sbox{\plotpoint}{\rule[-0.200pt]{0.400pt}{0.400pt}}%
\put(201.0,123.0){\rule[-0.200pt]{298.234pt}{0.400pt}}
\put(1439.0,123.0){\rule[-0.200pt]{0.400pt}{177.543pt}}
\put(201.0,860.0){\rule[-0.200pt]{298.234pt}{0.400pt}}
\put(201.0,123.0){\rule[-0.200pt]{0.400pt}{177.543pt}}
\end{picture}
  \caption{Averaged (non-interpolated) Precision/Recall curve}
  \label{fig:bestA}
\end{figure}
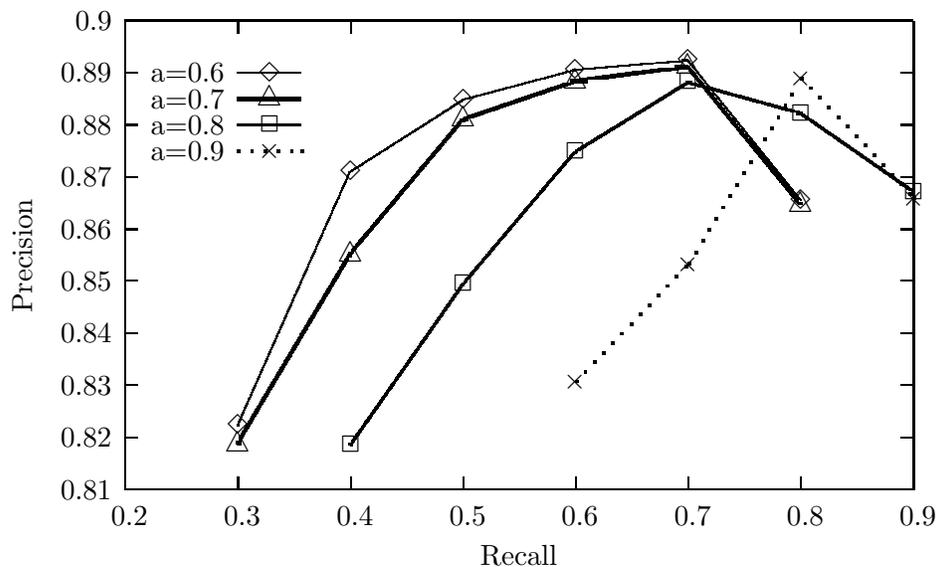

The obtained results are presented in the figure \ref{fig:bestA} and
the averages in table \ref{tab:avr-rez}.

\begin{table}[h]
  \begin{tabular}{|c|c|c|c|c|}
\hline
    \textbf{a}&\textbf{Accuracy}&\textbf{Precision}&\textbf{Recall}&\textbf{F-Measure}\\
\hline
0.6&0.93746&0.865545&0.763733&0.78103\\
\hline
0.7&0.940054&0.864756&0.779126&0.79017\\
\hline
0.8&0.948855&0.867199&0.837167&0.825773\\
\hline
0.9&0.960364&0.865917&0.911771&0.870373\\
\hline
\end{tabular}
\centering
  \caption{Average results}
  \label{tab:avr-rez}
\end{table}

With standard keyword extraction algorithm, there is a tradeoff
between the precision and the recall. When the recall augments because
there is more keywords extracted, the precision will fall because
there is more noise in the output. The curve \ref{fig:bestA} displays
a -- quite interesting -- different behaviour. The precision augments
in parallel with the recall up to a maximum -- around
$\mathrm{R(Prod,Ref) }= 0.7$ -- where the precision starts to fall.

This is explained by the goal of the algorithm. If we compare a cut
$\chi_1$ that contains a small set of concepts and a bigger cut
$\chi_2$, there is some chances that the latest contains more concepts
because it is lower in the hierarchy than $\chi_1$. That does not
always mean that $\chi_2$ is a more \emph{noisy} representation, but
it could also be that it is just a more precise set of concepts.

Then, if the recall is raising, it's explained by the fact that the
cut is lower in the hierarchy -- i.e nearer to the leaves -- and has
more chances to be near to the reference set of concept. At a point,
-- around $\mathrm{R(Prod,Ref) }= 0.7$ -- the cut is becoming more
precise than the reference annotation and the new concepts are
considered as \emph{noise}. That's why the precision starts to fall
for high values of recall.

A second observation can be made. For $a=0.6$ and $a=0.7$, the recall
is not defined over 0.9. The discussion at the end of the section
\ref{sec:eval:ext-scheme} explains this behaviour. In fact, for small
value of $a$, there is few chances that the extracted cut is specific
enough to have a good probability to match all the reference concepts,
and therefore it is hard to reach high value of recall.

\chapter{Future Work}
\label{cha:future-work}

As mentioned along this report, there is some part that could be
continued or added.

\section{Topic Extraction}
\label{sec:fw:topic-extraction}

\subsection*{Semantic Resource Quality}
\label{sec:fw:resource-quality}

One main problem with our topic extraction algorithm is presented in
the section \ref{sec:edr:version-issues} and
\ref{sec:eval:topic-extraction}. The resource used to construct the
semantic hierarchy -- which is the base for our computations -- and to
compute the similarity between concepts -- which is the base for our
evaluation metric -- does not seem to be adapted, by the lack of
information stored in the hierarchy, to the task pursued.  Adapting
the algorithm to use another semantic resource -- a newer version of
EDR or WordNet for example -- may perhaps help in getting a more
flexible tool.

In the same way, to render the evaluation independent, the semantic
similarity between the produced topics and the reference annotations
might be computed from a different semantic resource. This will help
getting results not biased by the fact that the hierarchy with which
we are evaluating may be erroneous. If the errors lead to a bad
extraction, at least, the evaluation will detect that the extraction
is erroneous.

\subsection*{Word Sense Disambiguation}
\label{sec:fw:wsd}

In the preprocessing step of the extraction algorithm, there may be
many concepts attached to one lemma in the document -- due to multiple
word senses. This problem could be resolved with two techniques:
\begin{itemize}
\item Word Sense Disambiguation, as explained in section
  \ref{sec:ext:disambiguation}.
\item In-line disambiguation, that would perform cut-off in the
  hierarchy to remove parts that are too far from the full hierarchy.
\end{itemize}

WSD could be performed at preprocessing, by using existing techniques
or developing new ones based on the semantic resource used for the
extraction. However, too much preprocessing can sometimes lead to loss
of precision -- i.e if the accuracy of the WSD task is 90\% and the
accuracy of the extraction task is also 90\%, then the system will
only perform at 80\%, which is not a good result.

In-line disambiguation have been discussed during the project, but the
lack of a good resource and of time discouraged us from developing a
good implementation. However, there are two main idea that can be used
to perform the word sense disambiguation during the extraction
process.

The first one will use some semantic similarity metric to remove
uninteresting sub-hierarchy from the initial one. If a sub-hierarhcy
seems to be too shallow -- it does not covers a lot of leaves concepts
-- or too far from the other side of the full hierarchy -- the
average semantic similarity between its concept and the other concepts
is too low --, it could be removed and a smaller hierarchy will be
used to search for cuts.

The second one will use a more complex scoring scheme for the concepts
(see section \ref{sec:concept-scoring}). In the actual scheme, it can
be unintuitive to forget about the relative frequency of each concept
in the document. It seems interesting to use information about how
present a concept is in the document to promote it. Unfortunately, it
is not obvious to know how to propagate the frequency of the leaves to
their super concepts. This scoring scheme should be explored in more
details in future projects.

\section{Segmentation}
\label{sec:fw:seg}

The segmentation algorithm used in this project is a really basic
implementation of the M.Hearst \cite{hearst94multiparagraph} text
tiling technique. Some interesting improvements could be made on
it.

The preprocessing task for this algorithm is really basic. Even if
M.Hearst has displayed results that show that such preprocessing was
sufficient, it could be interesting to study the impact of the
addition of a Part Of Speech tagger.

A second step would be the detection and the use in the algorithm of
the document structure. For example, the algorithm could create
windows (see section \ref{sec:seg:vector}) around full sentences
instead of simple words. In the same way, the paragraph segmentation
already provided by the document visual formating could be used to put
constraints on the window positions.

Our segmentation algorithm makes no use of the semantic resource
available. However, the algorithm could take advantage of the
conceptual representation constructed for the extraction algorithm to
deduce boundaries. For example, the vector used in the distance
computation could be filled with concepts instead of simple lemmas.
Then the distance between each conceptual vector will be computed
using one of the semantic distance techniques described in
\cite{budanitsky01semantic}.

\chapter{Conclusion}
\label{cha:conclusion}

In past systems, Automatic Topic Extraction was mainly treated in two
manners where \gloss[none]{statmeth}statistical methods and semantic
resources were needed. This project combines both techniques to
extract interesting topics from a document. A semantic hierarchy is
used to extract generic concepts summarizing the text themes. The
novel approach introduces the use of simple metrics that stay in the
control of the developer as they represent an intuitive manner of
choosing representative concepts in a complex ontology.

To reduce the computational bottle neck introduced by the use of a
conceptual hierarchy, the text is split in several segments from which
topics are extracted. The Text Tiling technique chosen gave good
results but revealed a lack of robustness linked to the difficulty of
deciding for a general setting of its main parameter: the window size.
To obtain the best results, user input is probably needed, but in this
project, the segmentation accuracy can stay approximative and one of
the high window sizes can be chosen to improve the algorithm
robustness.

The segmentation technique could get results improvements by including
a sentence detection algorithm to refine the boundaries placement. But
it must be kept in mind that this part of the project is to be
considered as a preprocessing step. As for most of the preprocessing
tasks, it is not always best to spend time refining it when the most
interesting part is the topic extraction algorithm.

The topic extraction algorithm is the main novelty in this project and
thus offered the possibility to examine interesting concepts and
emphasize their strength and weakness.  The construction of the
spanning \gloss[word]{DAG} over the segment's concepts is used to
deduce the context of each word and choose the best representative
concepts to describe the document's topics.

The novel evaluation method developed to measure extraction algorithm
accuracy gives the opportunity to observe interesting behaviours. The
evaluation displays good results in term of precision and recall.
However, the deficiencies in the semantic resource modified a bit
the expected algorithm behaviour. Because the semantic hierarchy is
too flat, the user does not have as much control on the extracted
concepts genericity as expected when the algorithm was constructed.

The project had the chance to develop interesting and novel parts of a
topic annotation system and pointed out important issues and their
supposed solutions. Finally two fully usable tools to segment texts
and perform topic extraction were implemented in the attempt to
contribute with an evolvable code.

This project proves that extracting concepts in place of simple
keywords can be beneficial and does not require too complex algorithm.
The loss of effectiveness was often coming from erroneous construction
of the semantic resource. Hence the development of this technique and
of the resources used could bring interesting extensions to existing
tools using keyword extraction algorithm.

Document indexing will gain flexibility if the returned documents are
conceptually identical to the query and do not simply contain the
queried keywords. Multilingual corpora alignment can also gain
robustness if they could be aligned on a conceptual basis.  However,
developing a better semantical topic extraction algorithm will need a
flawless linguistic resources and associated tools -- i.e.  tokenizer,
tagger -- that are fully compatible with it.

\appendix

\chapter*{Acknowledgements}
\thispagestyle{empty} I would like to acknowledge here Martin Rajman,
the project supervisor, who gave me important directions and interesting
inputs. Lonneke van der Plas, Florian Seydoux and Jean-Cedric
Chappelier for helping me on different parts of the project.

\addcontentsline{toc}{chapter}{Acknowledgements}

\listoffigures
\addcontentsline{toc}{chapter}{List Of Figures}

\nocite{*}

\renewcommand{\bibname}{References}
\bibliographystyle{alpha}
\bibliography{rapport}
\addcontentsline{toc}{chapter}{References}

\printgloss{glsbase,gloss}

\end{document}